\documentclass{ieeeaccess}

\usepackage{cite}
\usepackage{amsmath,amssymb,amsfonts}
\usepackage{graphicx}
\usepackage{textcomp}
\usepackage{url}
\usepackage{bm}
\usepackage{multirow}
\usepackage{booktabs}
\usepackage{makecell}
\usepackage{array}
\usepackage{pifont}
\usepackage{rotating}
\usepackage{subfig}
\usepackage{algorithmic}
\usepackage{placeins}

\graphicspath{{figures/}}

\usepackage{etoolbox}
\makeatletter
\providecommand{\@doi}{}
\providecommand{\@history}{}
\providecommand{\@pubid}{}
\patchcmd{\@maketitle}
  {{\vss\doifont Digital Object Identifier\space\@doi\vss}}
  {}
  {\typeout{PREPRINT: publisher DOI line removed.}}
  {\PackageWarning{preprint}{DOI patch not applied; check ieeeaccess.cls version}}
\def\headerlogo{}
\def\headerlogoall{}
\def\ps@titlepage{%
  \def\@oddhead{\parbox[t]{\textwidth}{\vspace*{-3mm}\hrulefill}}%
  \let\@evenhead\@oddhead
  \def\@oddfoot{\hfill{\footerpagefont\thepage}\hfill}%
  \let\@evenfoot\@oddfoot
}
\def\ps@headings{%
  \def\@oddhead{\parbox[t]{\textwidth}{\vspace*{-3mm}{\headerfont\rightmark}\par\hrulefill}}%
  \def\@evenhead{\parbox[t]{\textwidth}{\vspace*{-3mm}{\headerfont\leftmark}\par\hrulefill}}%
  \def\@oddfoot{\hfill{\footerpagefont\thepage}\hfill}%
  \let\@evenfoot\@oddfoot
}
\makeatother
\newlength{\myfigwidth}
\begin{document}


\title{Parameter-Efficient Multimodal Brain Tumor Segmentation With Stage-Wise Shared Adapters}

\author{\uppercase{Milad Rostamian}\authorrefmark{1},
\uppercase{Mohammadreza Mohammadi}\authorrefmark{1},
\uppercase{Ender Konukoglu}\authorrefmark{2}, and
\uppercase{Mohsen Soryani}\authorrefmark{1}}

\address[1]{School of Computer Engineering, Iran University of Science and Technology (IUST), Tehran, Iran (e-mail: milad\_rostamian@comp.iust.ac.ir; mrmohammadi@iust.ac.ir; soryani@iust.ac.ir)}
\address[2]{Computer Vision Laboratory, ETH Zurich, 8092 Zurich, Switzerland (e-mail: ender.konukoglu@vision.ee.ethz.ch)}

\markboth
{Rostamian \headeretal: Parameter-Efficient Multimodal Brain Tumor Segmentation}
{Rostamian \headeretal: Parameter-Efficient Multimodal Brain Tumor Segmentation}

\corresp{Corresponding author: Mohsen Soryani (e-mail: soryani@iust.ac.ir).}

\begin{abstract}
Different MRI modalities reveal different parts of a brain tumor. Segmentation models combine information from these modalities to outline tumor regions. Many models use a separate encoder for each modality. However, exchanging features among these encoders at several stages can increase model complexity. We introduce the Block-Wise Shared Adapter (BWSA), which combines compressed features from the modality encoders in a shared path and returns adapted features to each stream. In its fixed-blending form, BWSA adds the adapted features to the original features. Its trainable-gated variant (TG-BWSA) instead learns separate weights for the original and adapted features at each modality and encoder stage. We evaluate both variants on BraTS 2020 and BraTS 2023 with four segmentation backbones. On the official BraTS 2023 online validation set, BWSA increases average region-wise Dice by 1.3--7.6 percentage points relative to the corresponding unmodified backbones. TG-BWSA increases it by 2.8--7.7 points. Both variants add approximately 0.85--3.73\% parameters and reduce average HD95 relative to the unmodified backbones.
\end{abstract}


\titlepgskip=-21pt

\maketitle

\section{Introduction}
\label{sec:introduction}

Brain tumor segmentation from magnetic resonance imaging (MRI) supports diagnosis, treatment planning, and longitudinal assessment~\cite{Jyothi2023}. Multimodal protocols commonly include contrast-enhanced T1-weighted (T1ce), T1-weighted (T1), fluid-attenuated inversion recovery (FLAIR), and T2-weighted (T2) images, which provide complementary information about tumor and surrounding tissue~\cite{Baid2021BraTS}.

\begin{figure}[!tbh]
 \centering
 \includegraphics[width=\myfigwidth]{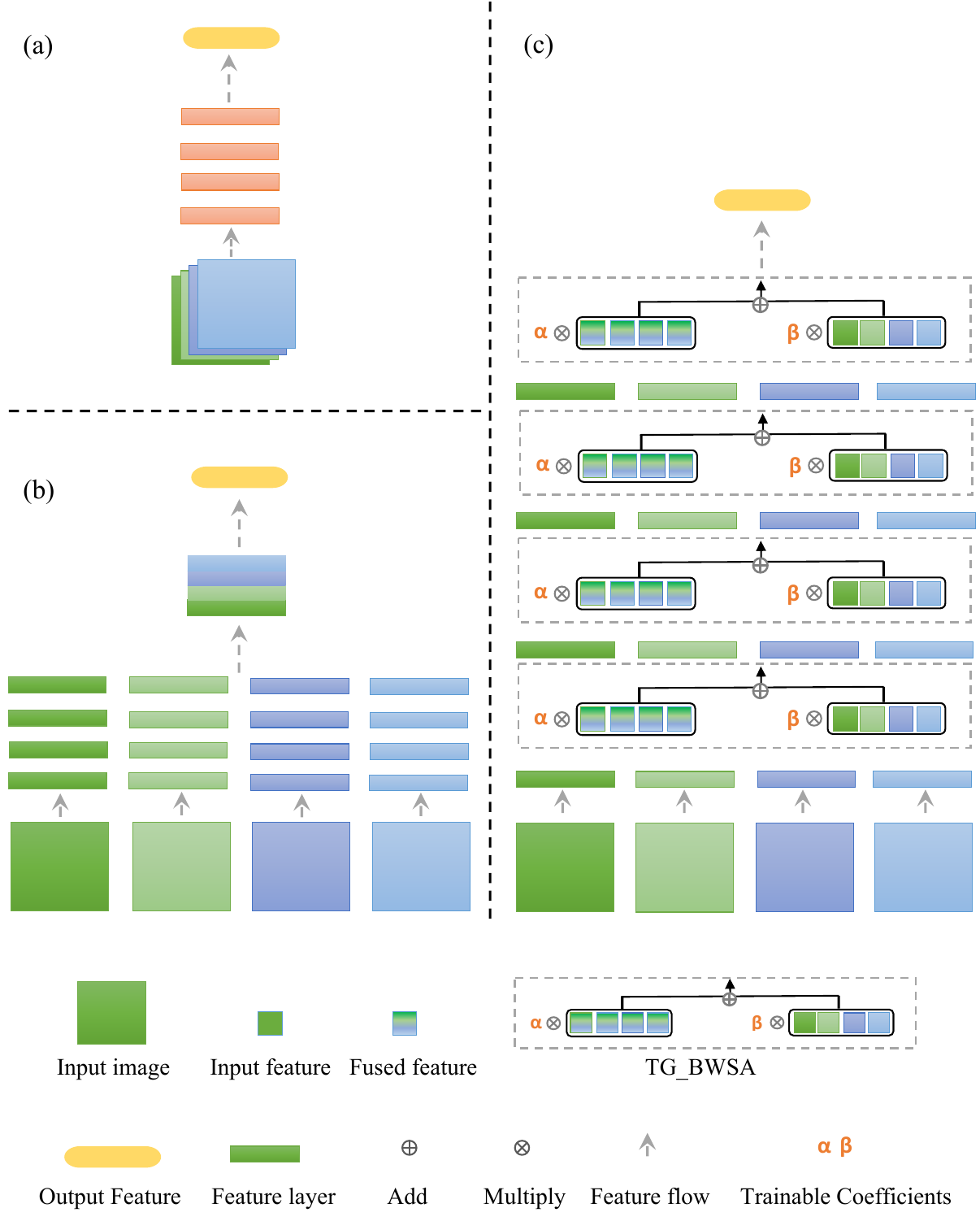}
 \caption{Representative fusion placements. (a) Early fusion concatenates the modalities at the input. (b) Late fusion combines independently produced predictions. (c) BWSA inserts a shared adaptation path between encoder stages. Its gated extension learns scalar weights for the original and adapted streams of each modality at each stage.}
 \label{fig:overview}
\end{figure}

Fusion strategies are commonly classified by where they combine the modalities~\cite{Stahlschmidt2022}. Early fusion concatenates modalities at the input, whereas late fusion merges independently produced outputs. Early fusion allows interaction throughout the network but does not preserve separate modality streams. Late fusion preserves those streams but postpones their interaction until the output. Intermediate fusion combines encoder features at one or more depths. Methods that use attention or gates can model cross-modal dependencies at these depths, but modality-specific branches and dense attention can increase parameter and computational costs~\cite{NestedFormer2023,Zhang2024,F2Net2024}.

Adapters have also been used in intermediate fusion. Shi et al. place fusion adapters at multiple layers of two pretrained Segment Anything Model (SAM) encoders to combine two MRI modality groups for glioma segmentation~\cite{Shi2025FusionSAM}. Lu et al. use a dual-stream adapter to align modality features before fusing them at multiple scales~\cite{Lu2025UMF}. We test whether a shared adapter can exchange features among four MRI encoders with a small increase in parameters. We also compare fixed and learned weights for combining the original and adapted features.

We introduce the \textbf{B}lock-\textbf{W}ise \textbf{S}hared \textbf{A}dapter (\textbf{BWSA}) for multimodal fusion between encoder stages (Fig.~\ref{fig:overview}c). BWSA combines the adapted and original features with fixed unit weights. Its gated extension, the \textbf{T}rainable \textbf{G}ated BWSA (\textbf{TG-BWSA}), learns two nonnegative scalars $(\alpha,\beta)$ for each modality and encoder stage. One weights the adapted feature and the other weights the original feature. These weights stay the same for every test image. Both forms use a single shared path rather than a full adapter for every modality. Separate expansion layers then return the shared features to the corresponding encoder streams.

We evaluate BWSA with fixed and learned stream weights on BraTS 2020 and BraTS 2023 using four CNN, Transformer, and hybrid backbones. We report local validation results and results from the BraTS 2023 online validation server. On online validation, fixed-weight BWSA raises average Dice by 1.3--7.6 percentage points relative to the unmodified backbones. The gated extension raises it by 2.8--7.7 percentage points. Both forms have nearly the same number of parameters because the gates add only two scalar weights per modality and stage. The reported overhead of 0.85--3.73\% compares the gated adapter with each unmodified backbone. We also evaluate RFNet~\cite{Ding2021} with both forms of BWSA under 14 incomplete input configurations and report the complete input result separately.

\noindent\textbf{Contributions.} The main contributions of this work are:
\begin{itemize}
    \item We introduce BWSA, a stage-wise shared adapter that can be inserted between encoder blocks in the four evaluated backbones.
    \item We extend BWSA to TG-BWSA by learning two scalar weights per modality and stage for the adapted multimodal stream and the original stream. The shared adapter path remains unchanged.
    \item We evaluate the shared adapter with fixed and learned weights. We also compare it with separate adapters for each modality.
    \item Missing modality experiments cover all 14 incomplete input configurations and report the complete input case separately.
\end{itemize}

\section{Related Work}

\subsection{Multimodal Fusion Methods}
Some models combine MRI sequences before feature extraction. Swin UNETR takes the four sequences as channels of one input and learns a joint representation~\cite{Hatamizadeh2022SwinUNETR}. This gives the encoder access to all sequences from its first layer, but does not keep a separate stream for each one. To retain modality-specific features, Nie et al. encode each modality separately and merge their high-level features before segmentation~\cite{Nie2016}. Fang et al. also use separate modality inputs and combine their features through hybrid attention, together with self-supervised feature learning~\cite{Fang2022}.

With separate streams, fusion can use learned interactions between their features. NestedFormer uses attention to combine high-level modality features and gating at lower scales~\cite{NestedFormer2023}. mmFormer uses an inter-modal Transformer to relate features from modality-specific encoders~\cite{Zhang2024}. F2Net maintains both modality-specific and shared representations, with modules that exchange information between them~\cite{F2Net2024}.

\subsection{Modality Weighting and Missing Inputs}
Keeping the modalities separate also allows a model to vary their contributions. RFNet learns which modality features matter in different tumor regions~\cite{Ding2021}. A2FSeg averages the available features and then uses attention to adjust their fusion~\cite{wang2023a2fseg}. These methods derive their weights from the input features. TG-BWSA instead learns one pair of stream weights for each modality and stage; the weights stay fixed across test inputs.

When a sequence is missing, the set of features available for fusion changes. HeMIS handles different subsets by averaging the embeddings of the available modalities~\cite{Havaei2016}. Chen et al. combine feature disentanglement and gated fusion to learn from incomplete inputs~\cite{chen2019robust}. Qiu et al. show that adding a modality can harm predictions in some regions and calibrate the fusion result accordingly~\cite{Qiu2025}.

\subsection{Adapters for Multimodal Fusion}
Adapters add a small trainable path to a model. Med-SA uses them to adapt pretrained SAM to medical image segmentation~\cite{Wu2024MedicalSAM}. Adapters can also take part in fusion. Shi et al. use fusion adapters between two pretrained SAM encoders to combine MRI modality groups at multiple layers~\cite{Shi2025FusionSAM}. Lu et al. use a dual-stream adapter to align modality features before multi-scale fusion~\cite{Lu2025UMF}. BWSA connects four separate modality streams through one shared adapter path at each encoder stage.

\section{Block-Wise Shared Adapter Framework}
\label{sec:method}

BWSA uses a shared path between encoder blocks to process compressed features from all modalities. It adds the adapted feature to the original feature of each modality. Its gated extension, TG-BWSA, learns separate nonnegative scalar weights for the original and adapted features at each modality and stage.

\subsection{Architectural Framework}
\label{sec:network_overview}

\begin{figure}[!tbh]
 \centering
 \includegraphics[width=\myfigwidth]{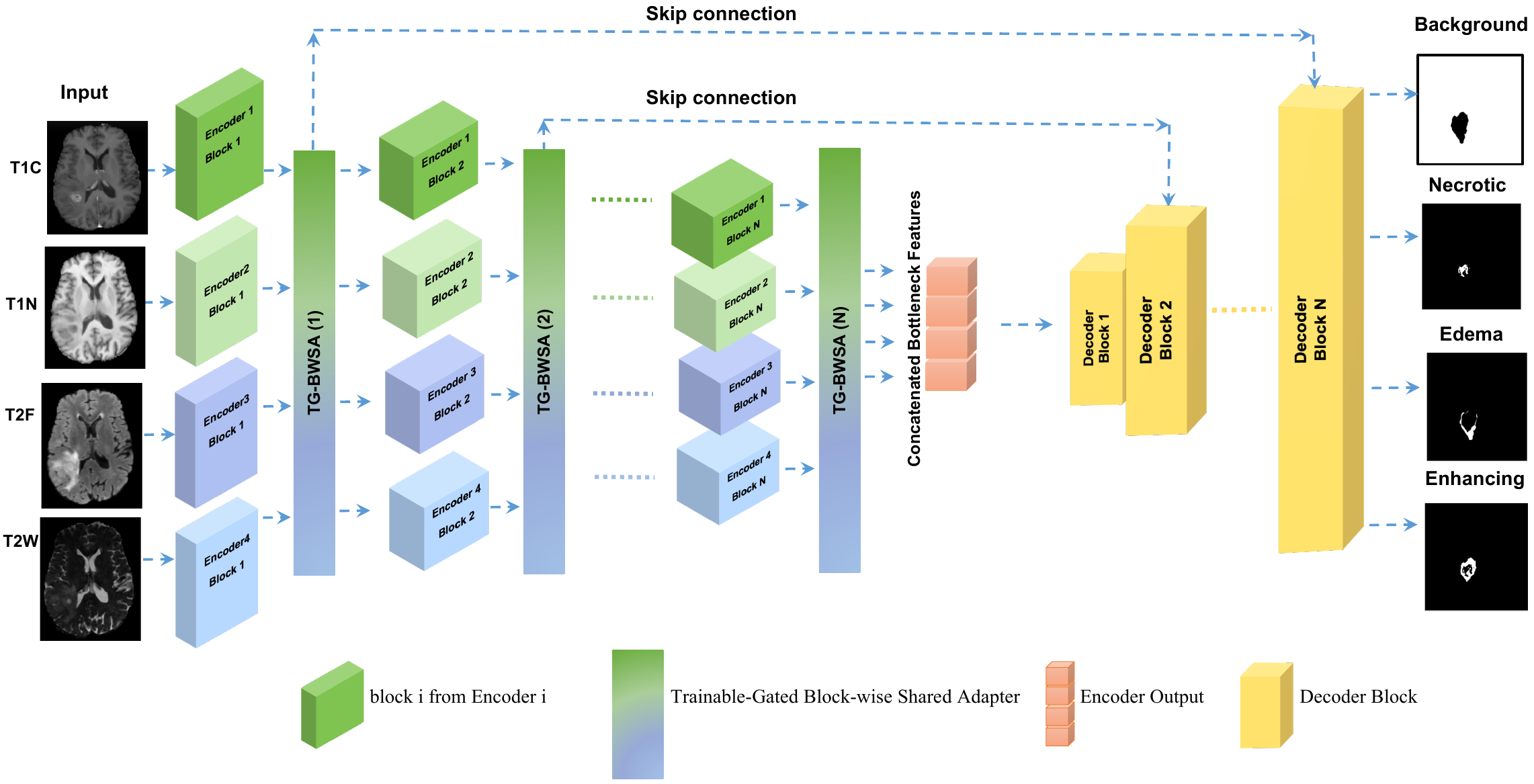}
 \caption{Overall multimodal framework. Each modality is processed by a separate encoder branch. The adapters connect successive encoder blocks, and their outputs are passed to a shared decoder through skip connections.}
 \label{fig:overall_architecture}
\end{figure}

We use a multi-encoder, single-decoder design (Fig.~\ref{fig:overall_architecture}). Let $M$ be the number of modalities. For BraTS, $M{=}4$ and the modalities are T1ce, T1, FLAIR, and T2. Each modality $m$ is processed by an encoder $E_m(\cdot)$ producing features at $L$ stages. At stage $l$, the modality-specific 3D feature maps are
\begin{equation}
\{X_m^{(l)}\}_{m=1}^{M}, \qquad X_m^{(l)} \in \mathbb{R}^{C \times D \times H \times W},
\end{equation}
where $C$ denotes the channel dimension and $(D,H,W)$ are depth, height, and width, respectively. For the 2D backbones, the depth dimension is omitted and the convolutions use the corresponding 2D kernels. An adapter transforms $\{X_m^{(l)}\}_{m=1}^{M}$ into $\{Y_m^{(l)}\}_{m=1}^{M}$, which are forwarded to stage $(l{+}1)$. At each stage, the updated modality features are concatenated to form the corresponding skip input to the shared decoder. The adapter changes the connections between encoder blocks but leaves the internal operations of those blocks unchanged.

\subsection{BWSA Module and Gated Extension}
\label{sec:Gated_Block-Wise_Adapter}

\begin{figure}[!t]
    \centering
    \subfloat[\textbf{TG-BWSA overview.}]{
        \includegraphics[width=\myfigwidth]{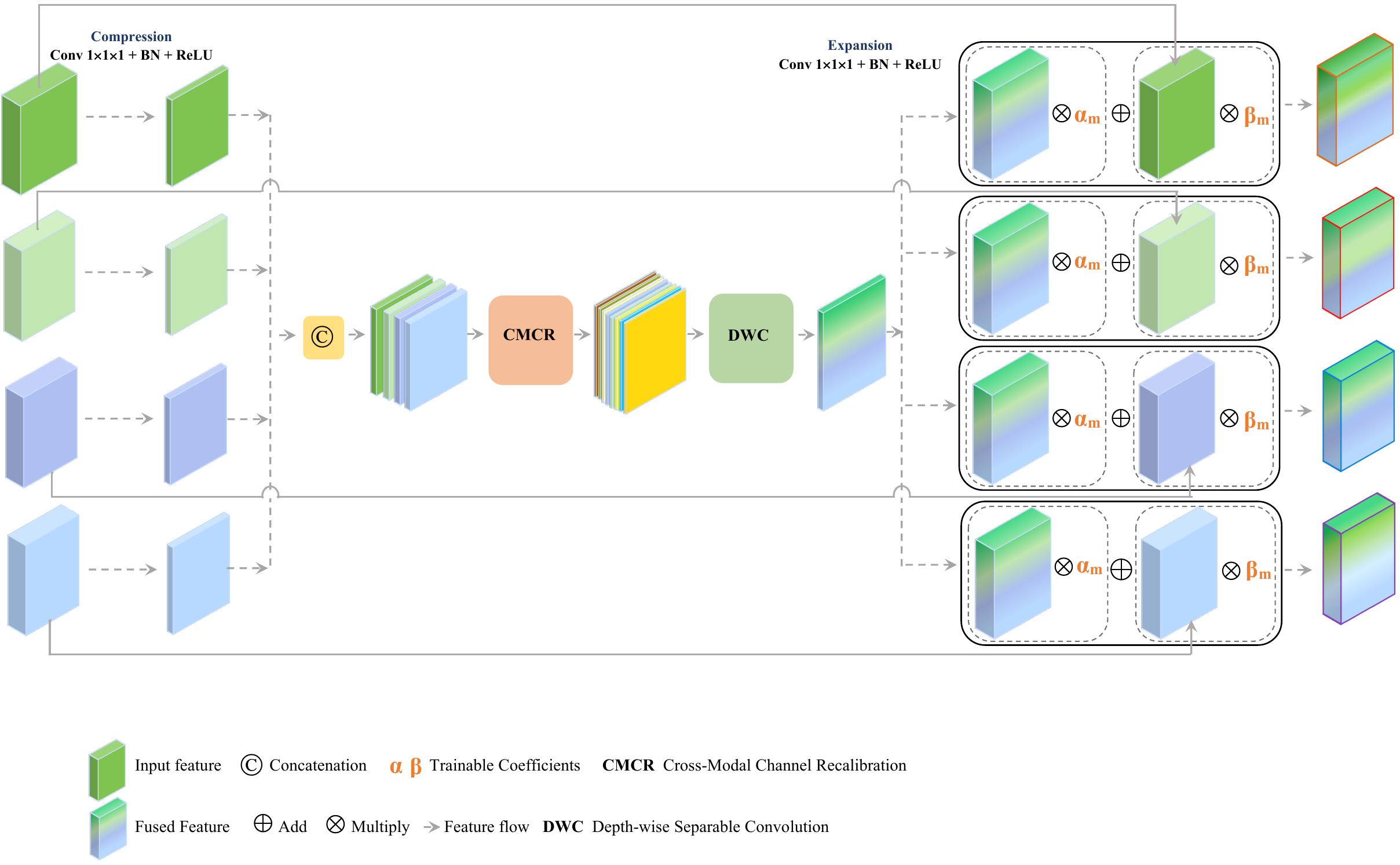}
        \label{fig:adapter_overview}
    }
    \vspace{0.25cm}
    \subfloat[\textbf{CMCR details.}]{
        \includegraphics[width=\myfigwidth]{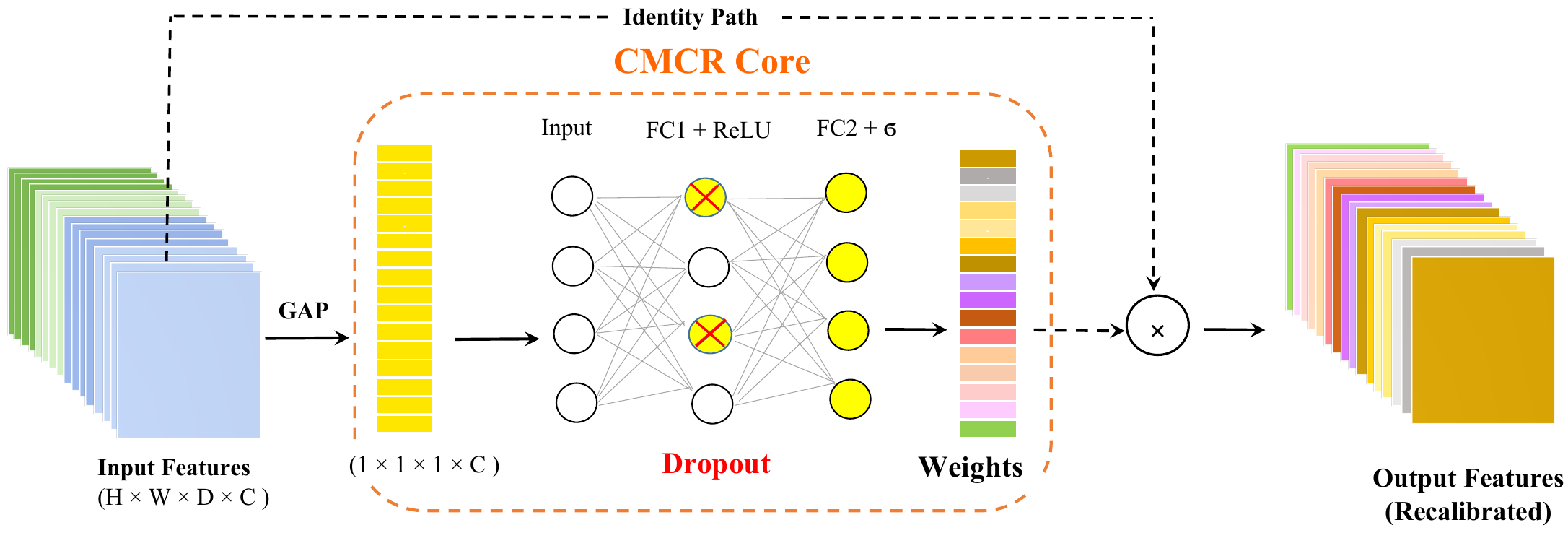}
        \label{fig:cmcr_detail}
    }
    \caption{\textbf{TG-BWSA.} (a) Compression $\rightarrow$ shared recalibration/refinement $\rightarrow$ gated redistribution. (b) CMCR operating on the concatenated multimodal latent tensor.}
    \label{fig:adapter_structure}
\end{figure}

Both variants use (i) channel compression and multimodal aggregation, (ii) shared cross-modal channel recalibration, (iii) spatial refinement, and (iv) residual blending. TG-BWSA adds learned weights in the final step.
Figure~\ref{fig:adapter_overview} illustrates the TG-BWSA overview, and Figure~\ref{fig:cmcr_detail} provides the CMCR details.

\subsubsection{Compression and Multimodal Aggregation}

Each modality feature map is first compressed to $C'=\lfloor rC\rfloor$ channels, with $r=0.5$, using a per-modality $1{\times}1{\times}1$ convolution:
\begin{equation}
X'^{(l)}_m
=
\mathrm{ReLU}\!\left(\mathrm{BN}\!\left(\mathrm{Conv}^{(m)}_{1 \times 1 \times 1}(X_m^{(l)})\right)\right)
\in \mathbb{R}^{C' \times D \times H \times W}.
\end{equation}
The compressed features are concatenated across modalities:
\begin{equation}
X_{\mathrm{cat}}^{\prime(l)} =
\left[ X'^{(l)}_1, \dots, X'^{(l)}_M \right]
\in \mathbb{R}^{MC' \times D \times H \times W}.
\end{equation}

\subsubsection{Cross-Modal Channel Recalibration (CMCR)}

To model cross-modal dependencies in the shared latent space, we apply a channel-attention (SE-style) recalibration~\cite{SandE2018Hu} to $X_{\mathrm{cat}}^{\prime(l)}$. The excitation path includes dropout with $p{=}0.1$:
\begin{equation}
\label{eq:cmcr}
\begin{aligned}
a^{(l)} &= \sigma\!\left(\mathbf{W}_2\,
\mathrm{Dropout}\!\left(\mathrm{ReLU}\!\left(
\mathbf{W}_1\,\mathrm{GAP}(X_{\mathrm{cat}}^{\prime(l)})
\right)\right)\right), \\
X''^{(l)} &= X_{\mathrm{cat}}^{\prime(l)} \odot a^{(l)}.
\end{aligned}
\end{equation}
where $a^{(l)}$ contains the channel weights, $\mathrm{GAP}(\cdot)$ is global average pooling, $\mathbf{W}_1,\mathbf{W}_2$ are fully-connected layers, $\sigma(\cdot)$ is sigmoid, and $\odot$ denotes channel-wise multiplication broadcast over $D{\times}H{\times}W$.

\subsubsection{Spatial Refinement and Expansion}

We apply a depthwise separable convolution to refine $X''^{(l)}$ spatially:
\begin{equation}
X'''^{(l)}
=
\mathrm{ReLU}\!\left(
\mathrm{BN}\!\left(
\mathrm{Conv}_{1\times1\times1}\bigl(
\mathrm{DWC}_{3\times3\times3}(X''^{(l)})
\bigr)\right)\right).
\end{equation}
Then, for each modality we project back to the original channel dimension $C$ to obtain an adapted feature map:
\begin{equation}
Z_m^{(l)}
=
\mathrm{ReLU}\!\left(
\mathrm{BN}\!\left(
\mathrm{Conv}^{(m)}_{1\times1\times1}(X'''^{(l)})
\right)\right)
\in \mathbb{R}^{C \times D \times H \times W}.
\end{equation}
The recalibration and refinement path up to $X'''^{(l)}$ is \emph{shared} across modalities. The final expansion to $Z_m^{(l)}$ is performed separately for each modality.

\subsubsection{Residual Blending and Trainable Gates}

BWSA adds the original and adapted features with unit weights:
\begin{equation}
\label{eq:base_scaling}
Y_{m,\mathrm{BWSA}}^{(l)}=X_m^{(l)}+Z_m^{(l)}.
\end{equation}
TG-BWSA instead learns two scalar gates per modality and stage:
\begin{equation}
\label{eq:scaling}
Y_m^{(l)}
=
\underbrace{(\beta_m^{(l)} + \epsilon)\,X_m^{(l)}}_{\text{Preservation}}
+
\underbrace{\alpha_m^{(l)}\,Z_m^{(l)}}_{\text{Adaptation}},
\qquad \epsilon = 10^{-3}.
\end{equation}
To enforce non-negativity, gates are parameterized by unconstrained variables and rectified via Softplus:
\begin{equation}
\alpha_m^{(l)}=\mathrm{Softplus}(\hat{\alpha}_m^{(l)}), \qquad
\beta_m^{(l)}=\mathrm{Softplus}(\hat{\beta}_m^{(l)}).
\end{equation}
TG-BWSA adds $2ML$ trainable scalar weights beyond BWSA, where $M$ is the number of modalities and $L$ is the number of adapter stages. With four modalities and four stages, this is $2\times4\times4=32$ weights. Both forms retain the same compression, channel-recalibration, spatial-refinement, and expansion layers.
We initialize $\hat{\alpha}_m^{(l)}$ and $\hat{\beta}_m^{(l)}$ such that $\alpha_m^{(l)}\approx 0.5$ and $\beta_m^{(l)}\approx 1.0$ after Softplus, giving greater initial weight to the original feature. We add an $\ell_2$ penalty to the gates at stages $2$ through $L$:
\begin{equation}
\label{eq:l2_gate}
\mathcal{L}_{\text{gate}}
=
\lambda
\sum_{l=2}^{L}\sum_{m=1}^{M}
\left((\alpha_m^{(l)})^{2} + (\beta_m^{(l)})^{2}\right),
\qquad \lambda=10^{-4}.
\end{equation}
During TG-BWSA training, the total objective is $\mathcal{L}_{\mathrm{seg}}+\mathcal{L}_{\mathrm{gate}}$, where $\mathcal{L}_{\mathrm{seg}}$ is the segmentation loss retained from the corresponding backbone. BWSA uses $\mathcal{L}_{\mathrm{seg}}$ without the gate penalty.

\begin{table}[t]
\caption{TG-BWSA forward pass at encoder stage $l$.}
\label{alg:gated_adapter}
\centering
\footnotesize
\begin{algorithmic}[1]
\STATE \textbf{Input:} modality features $\{X_m^{(l)}\}_{m=1}^{M}$
\STATE \textbf{Parameters:} compression ratio $r{=}0.5$ and $\epsilon{=}10^{-3}$
\FOR{$m=1$ to $M$}
\STATE $X_m^{\prime(l)} \leftarrow \mathrm{ReLU}(\mathrm{BN}(\mathrm{Conv}^{(m)}_{1\times1\times1}(X_m^{(l)})))$
\ENDFOR
\STATE $X_{\mathrm{cat}}^{\prime(l)} \leftarrow \mathrm{Concat}(X_1^{\prime(l)},\dots,X_M^{\prime(l)})$
\STATE $X''^{(l)} \leftarrow X_{\mathrm{cat}}^{\prime(l)} \odot \sigma(\mathbf{W}_2\,\mathrm{Dropout}(\mathrm{ReLU}(\mathbf{W}_1\,\mathrm{GAP}(X_{\mathrm{cat}}^{\prime(l)}))))$
\STATE $X'''^{(l)} \leftarrow \mathrm{ReLU}(\mathrm{BN}(\mathrm{Conv}_{1\times1\times1}(\mathrm{DWC}_{3\times3\times3}(X''^{(l)}))))$
\FOR{$m=1$ to $M$}
\STATE $Z_m^{(l)} \leftarrow \mathrm{ReLU}(\mathrm{BN}(\mathrm{Conv}^{(m)}_{1\times1\times1}(X'''^{(l)})))$
\STATE $\alpha_m^{(l)} \leftarrow \mathrm{Softplus}(\hat{\alpha}_m^{(l)})$
\STATE $\beta_m^{(l)} \leftarrow \mathrm{Softplus}(\hat{\beta}_m^{(l)})$
\STATE $Y_m^{(l)} \leftarrow (\beta_m^{(l)} + \epsilon)X_m^{(l)} + \alpha_m^{(l)}Z_m^{(l)}$
\ENDFOR
\STATE \textbf{return} $\{Y_m^{(l)}\}_{m=1}^{M}$
\end{algorithmic}
\end{table}

\paragraph*{Design choices.}
We use scalar gates to limit the number of added parameters. Channel-wise or spatial gates would provide finer control, but CMCR already weights channels in the combined features. Each learned gate applies to one modality and stage and remains fixed across test inputs. One shared path processes features from all modalities. Separate paths would repeat the same operations without combining their features inside the adapter. We compare the shared and separate paths in the PMA ablation under a similar parameter budget.

\section{Experimental Setup and Evaluation}
\label{sec:results}

\subsection{Datasets and Evaluation Metrics}

We conduct experiments on the BraTS 2020 (369 subjects) and BraTS 2023 (1,251 subjects) datasets~\cite{Menze2015,Bakas2017,Baid2021BraTS,BraTS2023Challenge}. Each dataset contains co-registered T1ce, T1, FLAIR, and T2 volumes with voxel-wise tumor annotations. We randomly assign 80\% of the subjects to training and 20\% to local validation. For BraTS 2023, we also report results on the official online validation set of 219 subjects. Dice is reported for all experiments. The local-validation Dice scores for 2D models are calculated on axial slices rather than reconstructed volumes and are not directly comparable with 3D local-validation or official online Dice scores. HD95 is also reported for the BraTS 2023 online validation set.

\noindent\textbf{BraTS online evaluation metrics:} We report \textbf{region-wise (voxel-wise)} Dice and HD95 for ET, TC, and WT. Lesion-wise results from the same online submission are reported in Table~\ref{tab:table6}.

\noindent\textbf{Data split and reproducibility:}
All train/validation splits are performed at the \emph{subject level} to prevent slice-level leakage. For each dataset, we randomly split subjects into 80\% training and 20\% validation once and keep this split fixed for all methods. Local-validation results are reported as mean$\pm$standard deviation over three independent runs with different random seeds for weight initialization and data order. Each run uses the same split and checkpoint-selection rule.

\subsection{Implementation Details}
\label{sec:implementation}

\noindent\textbf{Backbone Architectures:}
We integrated BWSA with fixed weights and with trainable gates into two 3D backbones, RFNet~\cite{Ding2021} and mmFormer~\cite{Zhang2024}, and two 2D backbones, a Multi-Encoder U-Net (MEUNet), implemented by us for this study, and Multi-Encoder FCT (MEFCT)~\cite{Tragakis2023}. F2Net~\cite{F2Net2024} uses Cross-modal Feature-enhanced Modules and Multimodal Collaboration Modules at multiple stages. We ran the authors' F2Net code and calculated Dice for its predictions using the same Dice calculation as for our 2D models. For the adapter comparisons, we retained the decoder and loss function of each backbone.

\noindent\textbf{Training Scenarios:}
We evaluated two training scenarios. RFNet, mmFormer, and MEUNet were trained from scratch. In the second scenario, the MEFCT encoders were initialized with self-supervised weights from RePre~\cite{Wang2022Repre}. The encoder freezing protocol for pretrained MEFCT, used by TG-BWSA, BWSA, and LoRA~\cite{hu2022lora} variants, is described below.

\noindent\textbf{3D Model Settings:}
For RFNet and mmFormer, inputs were z-score normalized volumetric patches. RFNet was trained for 300 epochs on $80 \times 80 \times 80$ patches with a batch size of~2 on a single NVIDIA A6000 GPU, while mmFormer was trained for 1000 epochs on $128 \times 128 \times 128$ patches with a batch size of~2 on the same hardware. Both used the Adam optimizer and a polynomial learning rate decay schedule. The checkpoint with the highest mean validation Dice was used for evaluation.

\noindent\textbf{2D Model Settings:}
MEUNet and MEFCT were trained for 100 epochs on min-max normalized $224 \times 224$ axial slices with a batch size of~16, distributed across four NVIDIA A6000 GPUs using data parallelism. The AdamW optimizer was employed, and the total loss combined weighted Categorical Cross-Entropy and Dice Loss.

\noindent\textbf{Self-supervised pretraining of FCT encoders:}
Following RePre~\cite{Wang2022Repre}, the four modality-specific encoders in MEFCT are pretrained \emph{individually} on unlabeled BraTS 2023 training images in the order T1ce, T1, FLAIR, and T2. For local validation, pretraining uses only images from the 80\% of subjects assigned to the training split. The held-out 20\% are excluded from both pretraining and supervised optimization. Each encoder jointly optimizes masked-patch inpainting and contrastive instance discrimination. We uniformly mask \(15\%\) of non-overlapping \(16\times16\) patches per slice and reconstruct them with a convolutional decoder using SSIM loss. A contrastive branch applies NT-Xent~\cite{Chen2020SimCLR} to projection heads over two independently augmented views. The total loss is \( \mathcal{L}_{\text{total}}=\lambda_{\text{SSIM}}\mathcal{L}_{\text{SSIM}}+\lambda_{\text{NT\text{-}Xent}}\mathcal{L}_{\text{contrast}} \), with \( \lambda_{\text{SSIM}}=0.8 \), \( \lambda_{\text{NT\text{-}Xent}}=0.2 \), and temperature \( \tau=0.2 \).

\noindent\textbf{Partial fine-tuning of pretrained MEFCT encoders:}
\label{subsec:frozen_ablation}
We use a two-phase \emph{partial fine-tuning} protocol for the pretrained MEFCT encoders. In Phase~1, all encoder layers are frozen. The decoder is trained for every variant, together with the BWSA, TG-BWSA, or LoRA modules when present. In Phase~2, the encoder bottleneck blocks are also unfrozen while the earlier encoder layers remain fixed. This protocol applies to the MEFCT variants in Table~\ref{tab:abl_mefct} and those marked with \textsuperscript{\dag} in the BraTS 2023 online validation table. Augmentations, losses, optimizer, scheduler, early stopping, and evaluation are held constant across the adapter variants.

\noindent\textbf{LoRA baseline (fairness and configuration):}
For MEFCT, we attach LoRA blocks after each self-attention block in the four encoders, including Blocks 1--4 and the bottleneck. Each LoRA module applies a low rank residual of the form $\tilde{x}=x+\frac{\alpha}{r}(xA)B$. We set $\alpha=2r$, use $r{=}16$ and $\alpha{=}32$ for the primary baseline, and apply dropout with $p{=}0.1$. The LoRA and TG-BWSA comparisons use the same data augmentation, losses, and optimization schedule. In both cases, the decoder and encoder bottlenecks are trainable. LoRA adds approximately $1.3\times10^{5}$ trainable parameters across the four encoders, whereas TG-BWSA adds approximately $9.0\times10^{5}$ parameters to MEFCT. The comparison therefore controls the training protocol but not the number or placement of added parameters.

\noindent\textbf{Per-modality adapters (PMA):}
PMA replicates the TG-BWSA structure (compression, CMCR, dual gates, refinement) independently for each modality, resulting in a comparable but slightly larger parameter budget than the shared TG-BWSA.

\subsection{Results}
\label{sec:results_main}
We report results for BWSA with fixed weights and its gated extension, TG-BWSA. The reported 0.85--3.73\% parameter increase is measured from each unmodified backbone to its TG-BWSA version. The evaluations include local full-modality validation, official BraTS 2023 online validation (Table~\ref{tab:table5}), and missing-modality inputs (Table~\ref{tab:missing_summary}). Section~\ref{sec:extended_results} reports parameter overhead and inference cost. Section~\ref{sec:ablation} reports component ablations, gate-initialization sensitivity, missing-modality experiments, and training analyses.

\subsubsection{Local Validation Results}

\noindent\textbf{Baselines and Notation:}
We evaluate BWSA with fixed stream weights and its gated extension. BWSA contains the shared CMCR and depthwise separable convolution path with $\alpha{=}1$ and $\beta{=}1$. TG-BWSA learns these two stream weights while retaining the shared path. The following notation is used in the tables.
\begin{itemize}
    \item Models augmented with the static \textbf{BWSA} are denoted with a single asterisk (*).
    \item Models using the gated extension, \textbf{TG-BWSA}, are denoted with a double asterisk (**).
    \item In detailed metric tables, \textbf{bold} marks the best value in each column, and \underline{underlining} marks the second-best.
\end{itemize}
BWSA with fixed weights increases mean Dice for RFNet, mmFormer, and MEUNet on both local validation sets, and the gated extension increases it further (Table~\ref{tab:local_summary}). Detailed region-wise results are reported in Tables~\ref{tab:table1}--\ref{tab:table4a}, and the pretrained MEFCT comparison is reported in Table~\ref{tab:abl_mefct}. MEUNet+TG-BWSA exceeds the F2Net result by 0.0230 Dice on BraTS 2020 (0.8974 vs.\ 0.8744) and by 0.0192 on BraTS 2023 (0.9240 vs.\ 0.9048).

\begin{table*}[!t]
\centering
\caption{Mean local full-modality Dice over WT, TC, and ET across three runs on BraTS 2020 and 2023. The 2D scores are slice-wise and are not directly comparable with the 3D scores. Bold marks the best variant within each backbone row. The last column shows the parameter increase from each unmodified backbone to its TG-BWSA version. Region-wise results and pretrained MEFCT comparisons appear in Tables~\ref{tab:table1}--\ref{tab:table4a} and Table~\ref{tab:abl_mefct}, respectively.}
\label{tab:local_summary}
\scriptsize
\setlength{\tabcolsep}{3pt}
\begin{tabular*}{\linewidth}{@{\extracolsep\fill}lccccccc}
\toprule
\multirow{2}{*}{\textbf{Backbone}} & \multicolumn{3}{c}{\textbf{BraTS 2020 mean Dice}} & \multicolumn{3}{c}{\textbf{BraTS 2023 mean Dice}} & \multirow{2}{*}{\textbf{TG-BWSA overhead}} \\
\cmidrule(lr){2-4} \cmidrule(lr){5-7}
& \textbf{Base} & \textbf{+BWSA} & \textbf{+TG-BWSA} & \textbf{Base} & \textbf{+BWSA} & \textbf{+TG-BWSA} & \\
\midrule
RFNet~\cite{Ding2021} & 0.8623 & 0.8693 & \textbf{0.8712} & 0.8941 & 0.8955 & \textbf{0.8982} & 3.73\% \\
mmFormer~\cite{Zhang2024} & 0.8635 & 0.8695 & \textbf{0.8715} & 0.8976 & 0.8995 & \textbf{0.9013} & 0.92\% \\
MEUNet & 0.8873 & 0.8957 & \textbf{0.8974} & 0.9169 & 0.9198 & \textbf{0.9240} & 2.94\% \\
F2Net~\cite{F2Net2024} & 0.8744 & --- & --- & 0.9048 & --- & --- & --- \\
\bottomrule
\end{tabular*}
\end{table*}

\noindent\textbf{Region-wise local-validation results:}
Tables~\ref{tab:table1}--\ref{tab:table4a} report the region-wise WT, TC, and ET results for the BraTS 2020 and BraTS 2023 local validation sets.

\begin{table}[!t]
\centering
\caption{BraTS 2020 local-validation Dice for 3D models (mean$\pm$standard deviation over three runs).}
\label{tab:table1}
\scriptsize
\setlength{\tabcolsep}{2pt}
\begin{tabular*}{\linewidth}{@{\extracolsep\fill}lcccc}
\toprule
\textbf{Model} & \textbf{WT} & \textbf{TC} & \textbf{ET} & \textbf{Mean} \\
\midrule
RFNet~\cite{Ding2021} & 0.9110{\scriptsize$\pm$0.0032} & 0.8825{\scriptsize$\pm$0.0031} & 0.7935{\scriptsize$\pm$0.0029} & 0.8623 \\
RFNet* & 0.9138{\scriptsize$\pm$0.0016} & 0.8847{\scriptsize$\pm$0.0019} & 0.8094{\scriptsize$\pm$0.0017} & 0.8693 \\
RFNet** & \underline{0.9161}{\scriptsize$\pm$0.0018} & \textbf{0.8861}{\scriptsize$\pm$0.0020} & \underline{0.8114}{\scriptsize$\pm$0.0011} & \underline{0.8712} \\
mmFormer~\cite{Zhang2024} & 0.9133{\scriptsize$\pm$0.0034} & 0.8798{\scriptsize$\pm$0.0030} & 0.7974{\scriptsize$\pm$0.0028} & 0.8635 \\
mmFormer* & 0.9151{\scriptsize$\pm$0.0028} & 0.8829{\scriptsize$\pm$0.0023} & 0.8105{\scriptsize$\pm$0.0026} & 0.8695 \\
mmFormer** & \textbf{0.9165}{\scriptsize$\pm$0.0017} & \underline{0.8847}{\scriptsize$\pm$0.0025} & \textbf{0.8134}{\scriptsize$\pm$0.0023} & \textbf{0.8715} \\
\bottomrule
\end{tabular*}
\end{table}

\begin{table}[!t]
\centering
\caption{BraTS 2020 slice-wise local-validation Dice for 2D models with trainable encoders (mean$\pm$standard deviation over three runs).}
\label{tab:table2a}
\scriptsize
\setlength{\tabcolsep}{2pt}
\begin{tabular*}{\linewidth}{@{\extracolsep\fill}lcccc}
\toprule
\textbf{Model} & \textbf{WT} & \textbf{TC} & \textbf{ET} & \textbf{Mean} \\
\midrule
MEUNet & 0.8724{\scriptsize$\pm$0.0029} & 0.8925{\scriptsize$\pm$0.0037} & 0.8971{\scriptsize$\pm$0.0031} & 0.8873 \\
MEUNet* & \underline{0.8838}{\scriptsize$\pm$0.0011} & 0.9019{\scriptsize$\pm$0.0015} & \underline{0.9014}{\scriptsize$\pm$0.0012} & \underline{0.8957} \\
MEUNet** & \textbf{0.8853}{\scriptsize$\pm$0.0013} & \textbf{0.9041}{\scriptsize$\pm$0.0011} & \textbf{0.9029}{\scriptsize$\pm$0.0010} & \textbf{0.8974} \\
F2Net~\cite{F2Net2024} & 0.8213{\scriptsize$\pm$0.0086} & \underline{0.9034}{\scriptsize$\pm$0.0035} & 0.8985{\scriptsize$\pm$0.0044} & 0.8744 \\
\bottomrule
\end{tabular*}
\end{table}

\begin{table}[!t]
\centering
\caption{BraTS 2023 local-validation Dice for 3D models (mean$\pm$standard deviation over three runs).}
\label{tab:table3}
\scriptsize
\setlength{\tabcolsep}{2pt}
\begin{tabular*}{\linewidth}{@{\extracolsep\fill}lcccc}
\toprule
\textbf{Model} & \textbf{WT} & \textbf{TC} & \textbf{ET} & \textbf{Mean} \\
\midrule
RFNet~\cite{Ding2021} & 0.9244{\scriptsize$\pm$0.0022} & 0.8976{\scriptsize$\pm$0.0023} & 0.8603{\scriptsize$\pm$0.0019} & 0.8941 \\
RFNet* & 0.9261{\scriptsize$\pm$0.0015} & 0.8990{\scriptsize$\pm$0.0002} & 0.8613{\scriptsize$\pm$0.0015} & 0.8955 \\
RFNet** & 0.9293{\scriptsize$\pm$0.0009} & 0.9007{\scriptsize$\pm$0.0011} & \underline{0.8647}{\scriptsize$\pm$0.0008} & 0.8982 \\
mmFormer~\cite{Zhang2024} & 0.9293{\scriptsize$\pm$0.0043} & 0.9011{\scriptsize$\pm$0.0040} & 0.8624{\scriptsize$\pm$0.0042} & 0.8976 \\
mmFormer* & \underline{0.9307}{\scriptsize$\pm$0.0017} & \underline{0.9029}{\scriptsize$\pm$0.0013} & \textbf{0.8648}{\scriptsize$\pm$0.0010} & \underline{0.8995} \\
mmFormer** & \textbf{0.9322}{\scriptsize$\pm$0.0015} & \textbf{0.9070}{\scriptsize$\pm$0.0016} & 0.8647{\scriptsize$\pm$0.0013} & \textbf{0.9013} \\
\bottomrule
\end{tabular*}
\end{table}

\begin{table}[!t]
\centering
\caption{BraTS 2023 slice-wise local-validation Dice for 2D MEUNet (mean$\pm$standard deviation over three runs).}
\label{tab:table4a}
\scriptsize
\setlength{\tabcolsep}{2pt}
\begin{tabular*}{\linewidth}{@{\extracolsep\fill}lcccc}
\toprule
\textbf{Model} & \textbf{WT} & \textbf{TC} & \textbf{ET} & \textbf{Mean} \\
\midrule
MEUNet & 0.8955{\scriptsize$\pm$0.0033} & 0.9291{\scriptsize$\pm$0.0030} & 0.9260{\scriptsize$\pm$0.0034} & 0.9169 \\
MEUNet* & \underline{0.9017}{\scriptsize$\pm$0.0012} & \underline{0.9306}{\scriptsize$\pm$0.0013} & \underline{0.9271}{\scriptsize$\pm$0.0010} & \underline{0.9198} \\
MEUNet** & \textbf{0.9037}{\scriptsize$\pm$0.0011} & \textbf{0.9345}{\scriptsize$\pm$0.0013} & \textbf{0.9338}{\scriptsize$\pm$0.0012} & \textbf{0.9240} \\
F2Net~\cite{F2Net2024} & 0.8634{\scriptsize$\pm$0.0249} & 0.9271{\scriptsize$\pm$0.0032} & 0.9240{\scriptsize$\pm$0.0024} & 0.9048 \\
\bottomrule
\end{tabular*}
\end{table}

\noindent\textbf{Adapter comparison on pretrained MEFCT:}
TG-BWSA has the highest mean Dice among the tested variants on BraTS 2020 and BraTS 2023 (Table~\ref{tab:abl_mefct}). On BraTS 2023, its mean Dice is 0.0036 higher than the reported LoRA baseline with $r{=}16$ (0.9095 vs.\ 0.9059). TG-BWSA adds approximately 0.9M parameters, compared with 0.127M for LoRA. The comparison uses the same training protocol, but the methods differ in parameter count and module placement.

\begin{table}[!tbh]
\centering
\caption{
Adapter comparison for 2D MEFCT on BraTS 2020 and 2023 local validation sets using slice-wise Dice under
\textbf{partial fine-tuning of pretrained encoders} (bottleneck blocks unfrozen).
Values are mean$\pm$standard deviation over three runs.
}
\label{tab:abl_mefct}
\scriptsize
\setlength{\tabcolsep}{2pt}
\begin{tabular*}{\linewidth}{@{\extracolsep\fill}lcccc}
\toprule
\multicolumn{5}{c}{\textbf{(a) BraTS 2020}} \\ \midrule
\textbf{Model} & \textbf{WT} & \textbf{TC} & \textbf{ET} & \textbf{Mean} \\
\midrule
MEFCT & 0.8614{\scriptsize$\pm$0.0041} & 0.8792{\scriptsize$\pm$0.0035} & 0.8775{\scriptsize$\pm$0.0047} & 0.8727 \\
MEFCT* & \underline{0.8708}{\scriptsize$\pm$0.0013} & 0.8802{\scriptsize$\pm$0.0010} & \underline{0.8894}{\scriptsize$\pm$0.0014} & \underline{0.8801} \\
MEFCT** & \textbf{0.8720}{\scriptsize$\pm$0.0011} & \textbf{0.8871}{\scriptsize$\pm$0.0009} & \textbf{0.8911}{\scriptsize$\pm$0.0008} & \textbf{0.8834} \\
MEFCT-LoRA~\cite{hu2022lora} & 0.8694{\scriptsize$\pm$0.0014} & \underline{0.8813}{\scriptsize$\pm$0.0019} & 0.8873{\scriptsize$\pm$0.0007} & 0.8793 \\
\bottomrule
\end{tabular*}

\vspace{1ex} 

\begin{tabular*}{\linewidth}{@{\extracolsep\fill}lcccc}
\toprule
\multicolumn{5}{c}{\textbf{(b) BraTS 2023}} \\ \midrule
\textbf{Model} & \textbf{WT} & \textbf{TC} & \textbf{ET} & \textbf{Mean} \\
\midrule
MEFCT & 0.8741{\scriptsize$\pm$0.0044} & 0.9152{\scriptsize$\pm$0.0037} & 0.9080{\scriptsize$\pm$0.0042} & 0.8991 \\
MEFCT* & \underline{0.8879}{\scriptsize$\pm$0.0013} & 0.9183{\scriptsize$\pm$0.0009} & \underline{0.9144}{\scriptsize$\pm$0.0014} & \underline{0.9069} \\
MEFCT** & \textbf{0.8902}{\scriptsize$\pm$0.0011} & \textbf{0.9223}{\scriptsize$\pm$0.0008} & \textbf{0.9160}{\scriptsize$\pm$0.0010} & \textbf{0.9095} \\
MEFCT-LoRA~\cite{hu2022lora} & 0.8859{\scriptsize$\pm$0.0015} & \underline{0.9190}{\scriptsize$\pm$0.0008} & 0.9129{\scriptsize$\pm$0.0016} & 0.9059 \\
\bottomrule
\end{tabular*}
\end{table}

\subsubsection{Region-wise Online Validation}

\noindent\textbf{BraTS 2023 online validation protocol:}
For each model, we trained three runs on the same 80\% subject-level training split. We selected the run with the highest mean Dice on the held-out 20\% local validation split and submitted predictions from that run to the official BraTS 2023 online validation server. Neither local nor online validation subjects were used for training.

On the official BraTS 2023 online validation set (Table~\ref{tab:table5}), RFNet has an average Dice of 0.8314 without an adapter. BWSA raises it to 0.8639, and learning the weights raises it to 0.8654. The corresponding average HD95 values are 28.80, 14.37, and 16.34\,mm. Thus, fixed weights give RFNet a lower average HD95 than learned weights. With mmFormer, MEUNet, and MEFCT, the gated form increases average Dice over the unmodified backbone by 0.0769, 0.0283, and 0.0450, respectively. It also reduces average HD95 for these three backbones. The added effect of learned weights varies by region and backbone.

\begin{table*}[!t]
\centering
\caption{BraTS 2023 online validation results for all models (Dice/HD95). \textsuperscript{\dag}Pretrained encoders use partial fine-tuning according to the protocol in Sec.~\ref{subsec:frozen_ablation}.}
\label{tab:table5}
\scriptsize
\setlength{\tabcolsep}{2.5pt}
\begin{tabular*}{\linewidth}{@{\extracolsep\fill}lcccccccc}
\toprule
\textbf{Model} & \textbf{Dice ET} & \textbf{Dice TC} & \textbf{Dice WT} & \textbf{HD95 ET} & \textbf{HD95 TC} & \textbf{HD95 WT} & \textbf{Dice Avg} & \textbf{HD95 Avg} \\
\midrule
RFNet~\cite{Ding2021} & 0.7889 & 0.8201 & 0.8853 & 35.4500 & 26.8100 & 24.1500 & 0.8314 & 28.8033 \\
RFNet + BWSA & \underline{0.8313} & \textbf{0.8511} & \underline{0.9093} & 20.5500 & 16.3800 & \textbf{6.1700} & \underline{0.8639} & 14.3667 \\
RFNet + TG-BWSA & \textbf{0.8364} & \underline{0.8472} & \textbf{0.9127} & \underline{18.9200} & 23.3400 & \underline{6.7700} & \textbf{0.8654} & 16.3433 \\
mmFormer~\cite{Zhang2024} & 0.6209 & 0.7025 & 0.8289 & 48.1700 & 33.2500 & 24.7900 & 0.7174 & 35.4033 \\
mmFormer + BWSA & 0.7502 & 0.7745 & 0.8563 & 26.9400 & 28.6200 & 16.5000 & 0.7937 & 24.0200 \\
mmFormer + TG-BWSA & 0.7685 & 0.7751 & 0.8394 & 23.5300 & 22.4800 & 19.8300 & 0.7943 & 21.9467 \\
Multi-Enc U-Net & 0.7467 & 0.8147 & 0.8918 & 26.4400 & 16.9600 & 9.1200 & 0.8177 & 17.5067 \\
Multi-Enc U-Net + BWSA & 0.7607 & 0.8329 & 0.8972 & 22.5500 & \underline{13.2300} & 7.0100 & 0.8303 & \underline{14.2633} \\
Multi-Enc U-Net + TG-BWSA & 0.7962 & 0.8378 & 0.9040 & \textbf{18.0200} & \textbf{12.5800} & 7.8800 & 0.8460 & \textbf{12.8267} \\
Multi-Enc FCT\textsuperscript{\dag} & 0.7160 & 0.7533 & 0.8691 & 31.7300 & 29.9700 & 9.9900 & 0.7795 & 23.8967 \\
Multi-Enc FCT + BWSA\textsuperscript{\dag} & 0.7652 & 0.8120 & 0.8856 & 18.5655 & 18.8615 & 7.9558 & 0.8209 & 15.1276 \\
Multi-Enc FCT + TG-BWSA\textsuperscript{\dag} & 0.7732 & 0.8133 & 0.8871 & 18.7600 & 18.1300 & 7.7400 & 0.8245 & 14.8767 \\
Multi-Enc FCT + LoRA\textsuperscript{\dag}~\cite{hu2022lora} & 0.7692 & 0.8024 & 0.8847 & 20.2674 & 24.2854 & 7.7655 & 0.8188 & 17.4394 \\
\bottomrule
\end{tabular*}
\end{table*}

\subsubsection{Lesion-wise Online Validation}
The BraTS 2023 segmentation evaluation reports both region-wise and lesion-wise Dice and HD95. The lesion-wise metrics, which are used for challenge ranking, evaluate connected lesions and penalize missed and false-positive components~\cite{BraTS2023Metrics,BraTS2023Ranking}. Table~\ref{tab:table5} reports region-wise results for the evaluated models. Table~\ref{tab:table6} includes lesion-wise results from the same RFNet+BWSA and RFNet+TG-BWSA online submissions, alongside published comparators. Region-wise Dice measures voxel overlap within each tumor region. We therefore report the two types separately.

\begin{table*}[!t]
\centering
\caption{BraTS 2023 online validation lesion-wise Dice and HD95. Comparator values and anonymized team identifiers are transcribed from Zheng et al.~\cite{Zheng2025}.}
\label{tab:table6}
\scriptsize
\setlength{\tabcolsep}{2.5pt}
\begin{tabular*}{\linewidth}{@{\extracolsep\fill}lcccccccc}
\toprule
\textbf{Model} & \makecell{\textbf{L-Dice}\\\textbf{ET}} & \makecell{\textbf{L-Dice}\\\textbf{TC}} & \makecell{\textbf{L-Dice}\\\textbf{WT}} & \makecell{\textbf{L-HD95}\\\textbf{ET}} & \makecell{\textbf{L-HD95}\\\textbf{TC}} & \makecell{\textbf{L-HD95}\\\textbf{WT}} & \makecell{\textbf{Avg}\\\textbf{L-Dice}} & \makecell{\textbf{Avg}\\\textbf{L-HD95}} \\
\midrule
\textbf{Team @s7zuqer} & \textbf{0.866} & \textbf{0.89} & \textbf{0.909} & \textbf{10.594} & \textbf{7.228} & \underline{11.086} & \textbf{0.888} & \textbf{9.636} \\
\textbf{Team @simis} & \underline{0.857} & \underline{0.876} & \underline{0.908} & \underline{13.649} & \underline{12.825} & 11.270 & \underline{0.880} & \underline{12.581} \\
\textbf{Team @qian8} & 0.856 & 0.868 & 0.904 & 14.070 & 13.778 & \textbf{10.735} & 0.876 & 12.861 \\
Faking\_it~\cite{ferreira2024faking} & 0.854 & 0.858 & 0.887 & 17.321 & 18.007 & 19.322 & 0.866 & 18.217 \\
BiomedMBZ~\cite{maani2024advanced} & 0.837 & 0.859 & 0.903 & 25.451 & 19.185 & 12.685 & 0.866 & 19.107 \\
Shenhai Zheng~\cite{Zheng2025} & 0.823 & 0.860 & 0.862 & 25.602 & 15.110 & 29.291 & 0.848 & 23.334 \\
\textbf{Ours (RFNet + BWSA)} & 0.834 & 0.828 & 0.883 & 21.585 & 31.274 & 17.343 & 0.848 & 23.401 \\	
\textbf{Ours (RFNet + TG-BWSA)} & 0.832 & 0.837 & 0.888 & 21.309 & 25.153 & 15.369 & 0.852 & 20.610 \\
\bottomrule
\end{tabular*}
\end{table*}

\noindent\textbf{Lesion-wise comparison:}
RFNet+TG-BWSA has an average lesion-wise Dice of 0.852 and average lesion-wise HD95 of 20.61, compared with 0.848 and 23.33 for the method of Zheng et al.~\cite{Zheng2025}. The three highest-scoring listed teams have average lesion-wise Dice values of 0.876--0.888, above the 0.852 obtained by RFNet+TG-BWSA. The listed teams used different training and inference pipelines. Their scores show how our results compare with other submissions, but they do not isolate the effect of the adapter.

\subsubsection{Qualitative Analysis}
\begin{figure*}[!tbhp]
\centering
\subfloat[2D models (MEFCT and MEUNet)]{%
    \includegraphics[height=.51\textwidth]{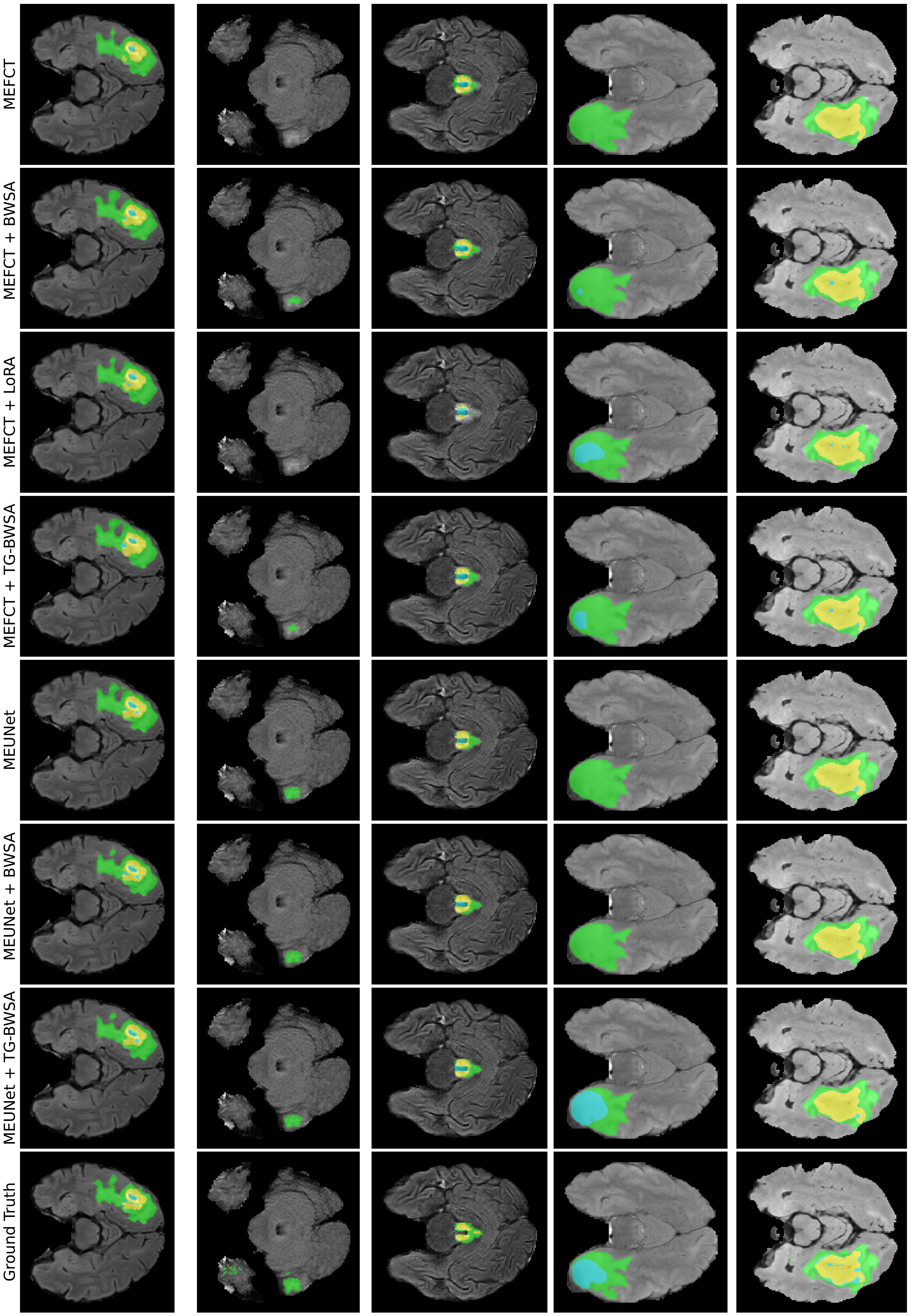}
    \label{fig:qualitative1}
}
\hspace{0.02\textwidth}
\subfloat[3D models (RFNet and mmFormer)]{%
    \includegraphics[height=.51\textwidth]{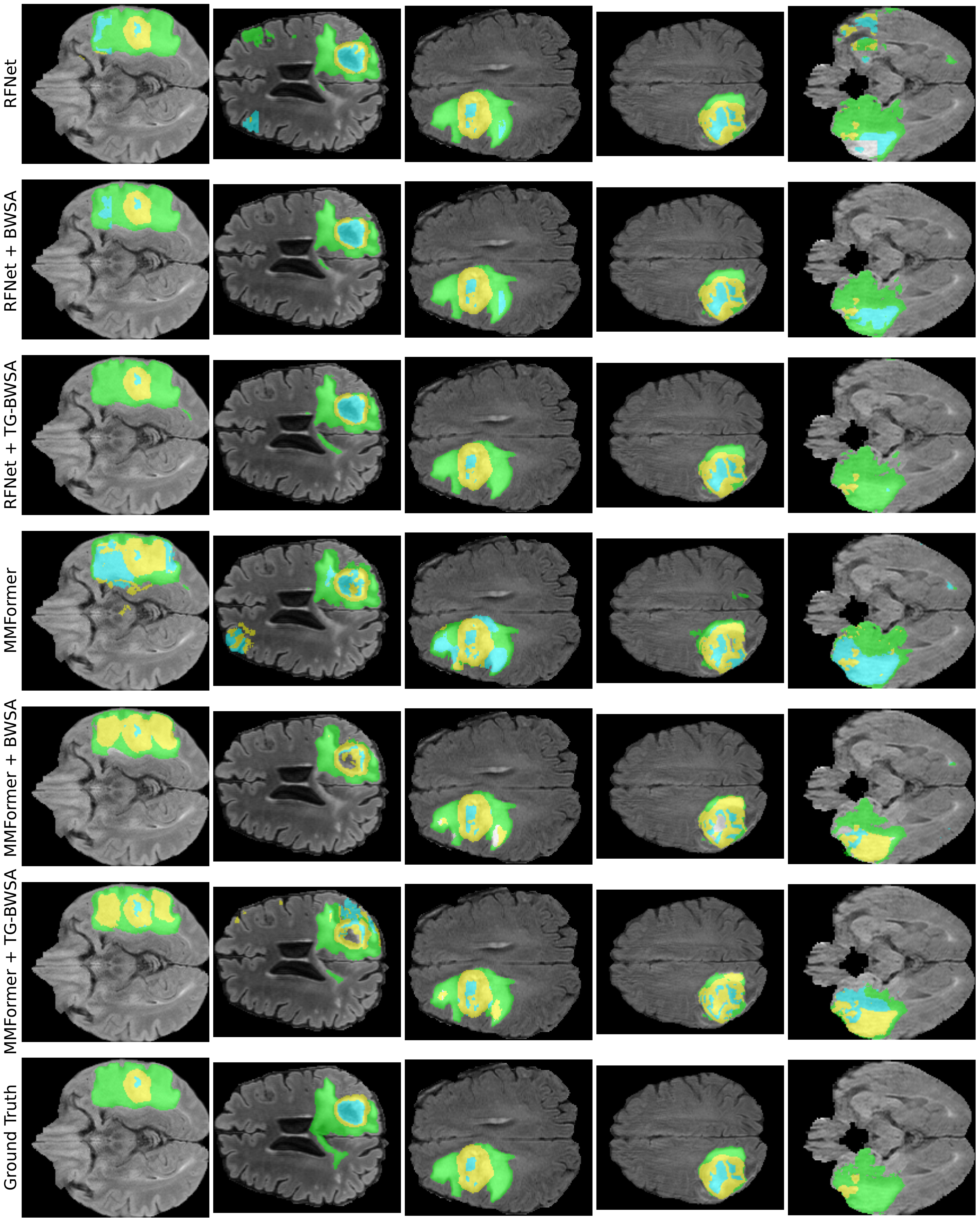}
    \label{fig:qualitative2}
}
\caption{Qualitative segmentation comparisons on randomly sampled BraTS 2023 local validation cases. Predictions are shown for the unmodified backbones and their adapter variants. Cyan denotes non-enhancing tumor, green denotes edema, and yellow denotes enhancing tumor.}
\label{fig:qualitative_combined_subfig}
\end{figure*}

Figure~\ref{fig:qualitative_combined_subfig} shows cases sampled at random from the BraTS 2023 local validation set for the 2D and 3D backbones, without ranking cases by model performance. In the displayed 2D cases, the adapter variants recover parts of the annotated tumor extent that are absent or undersegmented in some backbone predictions. TG-BWSA also yields boundaries that more closely follow the ground truth in several examples. In the displayed 3D cases, BWSA and TG-BWSA reduce several isolated false positive regions and correct some areas of oversegmentation or undersegmentation relative to the corresponding backbones.

\subsection{Parameter Overhead and Inference Cost}
\label{sec:extended_results}

\subsubsection{Parameter Overhead}
Table~\ref{tab:param_efficiency} compares each unmodified backbone with its TG-BWSA version. BWSA uses the same adapter layers and lacks only the $2ML$ scalar gates, so its parameter count is nearly identical to that of TG-BWSA. The percentage overhead in the table is relative to the unmodified backbone and is not the cost of gating alone.

\begin{table}[!tbh]
\centering
\caption{Parameter counts (millions) for each unmodified backbone and its TG-BWSA version. The reported overhead compares these two models. Counts are rounded to one decimal place; percentages may differ from calculations using the displayed counts.}
\label{tab:param_efficiency}
\begin{tabular*}{\linewidth}{@{\extracolsep\fill}lccc}
\toprule
\textbf{Backbone} & $P_{\text{base}}$ & $P_{\text{+TG-BWSA}}$ & $\Delta_{\%}$ \\
\midrule
RFNet & 9.0  & 9.3  & 3.73 \\
mmFormer & 36.7 & 37.0 & 0.92 \\
Multi-Enc. U-Net & 29.8 & 30.7 & 2.94 \\
Multi-Enc. FCT & 102.7 & 103.6 & 0.85 \\
\bottomrule
\end{tabular*}
\end{table}

\subsubsection{Inference Cost}
We measured wall-clock latency for a single forward pass with an $80^3$ input, peak GPU memory, and theoretical GFLOPs for RFNet with and without TG-BWSA on one NVIDIA A6000 GPU. Table~\ref{tab:inference_efficiency} shows increases of 2.4\% (37~ms) in latency, 8.2\% in peak memory, and 4.1\% in GFLOPs. These measurements apply to the reported input size, backbone, implementation, and hardware configuration.

\begin{table}[tbhp!]
\centering
\caption{Inference cost for one forward pass with an $80^3$ input on an NVIDIA A6000 GPU.}
\label{tab:inference_efficiency}
\begin{tabular*}{\linewidth}{@{\extracolsep\fill}lccc}
\toprule
\textbf{Model} & \textbf{Latency (s/input)} & \textbf{GPU Memory (GB)} & \textbf{GFLOPs} \\
\midrule
RFNet~\cite{Ding2021} & 1.544 & 1.95 & 102.8 \\
RFNet** & 1.581 & 2.11 & 107.0 \\
\bottomrule
\end{tabular*}
\end{table}

\subsection{Ablation Studies and Training Analysis}
\label{sec:ablation}

\begin{figure}[!tbh]
\centering
\includegraphics[width=\myfigwidth]{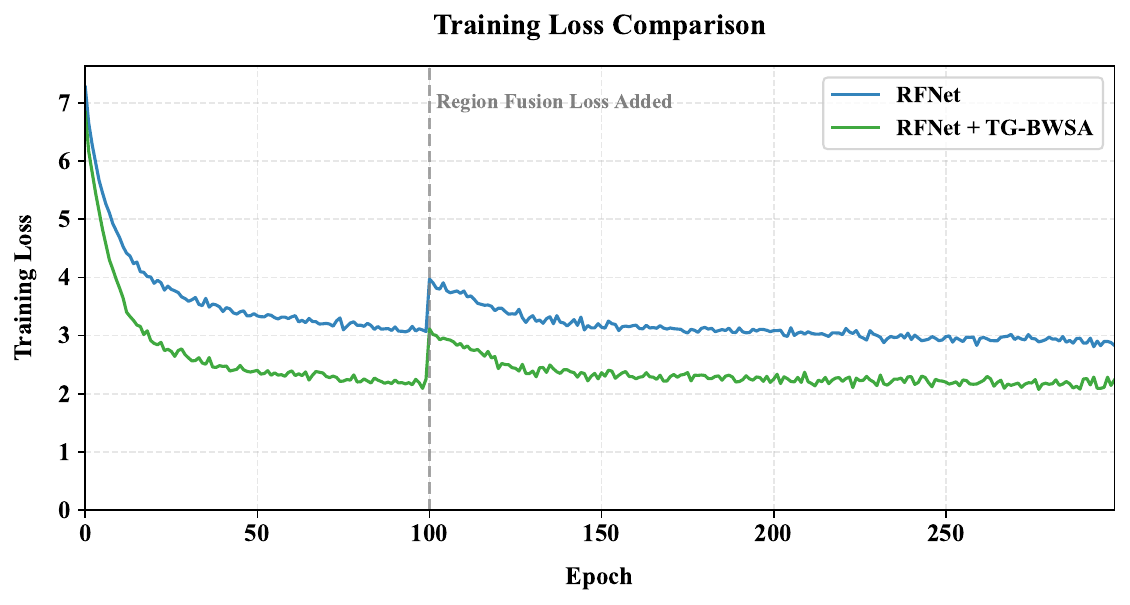}
\caption{
Training loss for RFNet and RFNet+TG-BWSA. The dashed line marks activation of the region-fusion loss near epoch 100. The curves describe optimization on the training data and should be interpreted together with the validation results.}
\label{fig:training_loss}
\end{figure}

\begin{figure*}[!htbp]
  \centering
  \subfloat[Gate Trajectories During Training (Encoder Stages 1-4)\label{fig:gate_trajectories}]{%
    \begin{tabular}{cc}
      \includegraphics[width=0.45\textwidth]{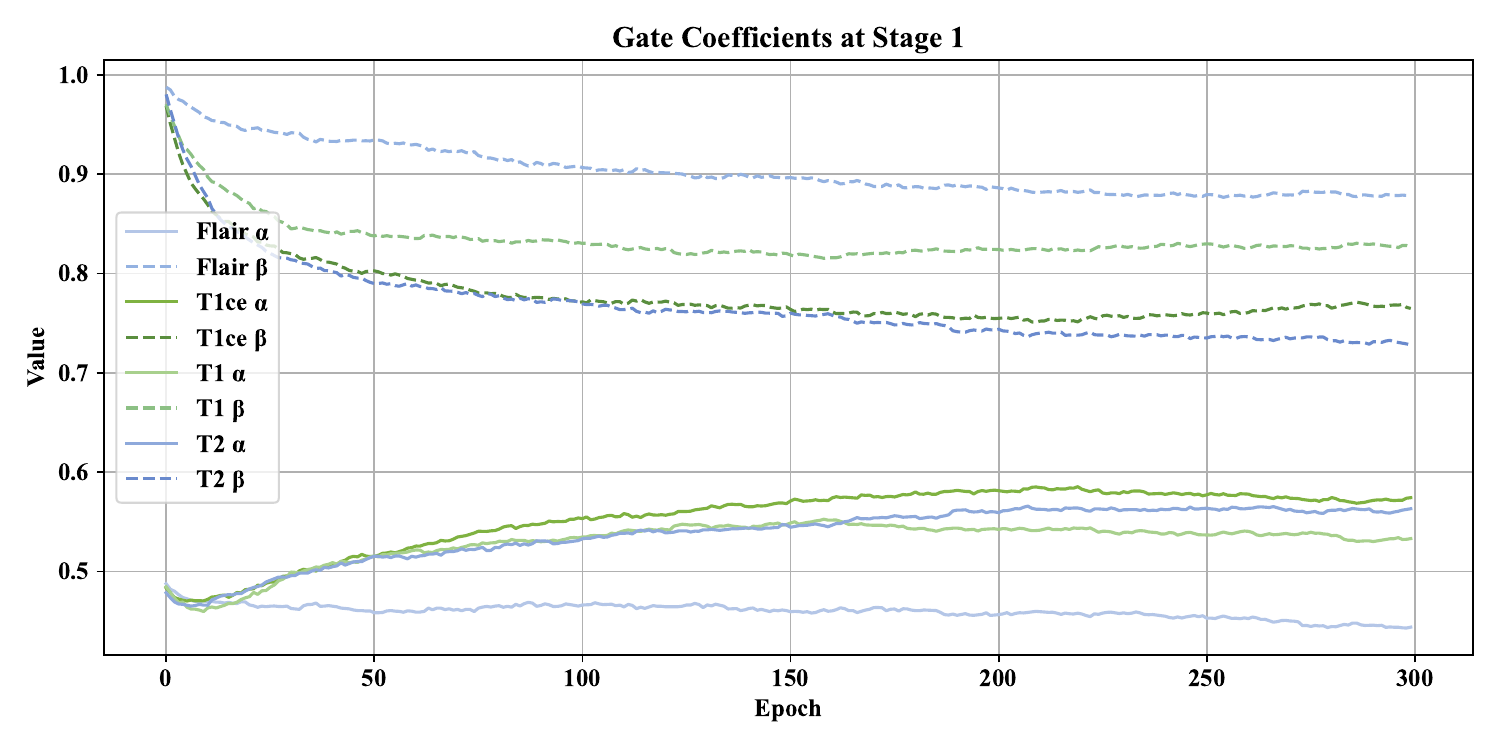} &
      \includegraphics[width=0.45\textwidth]{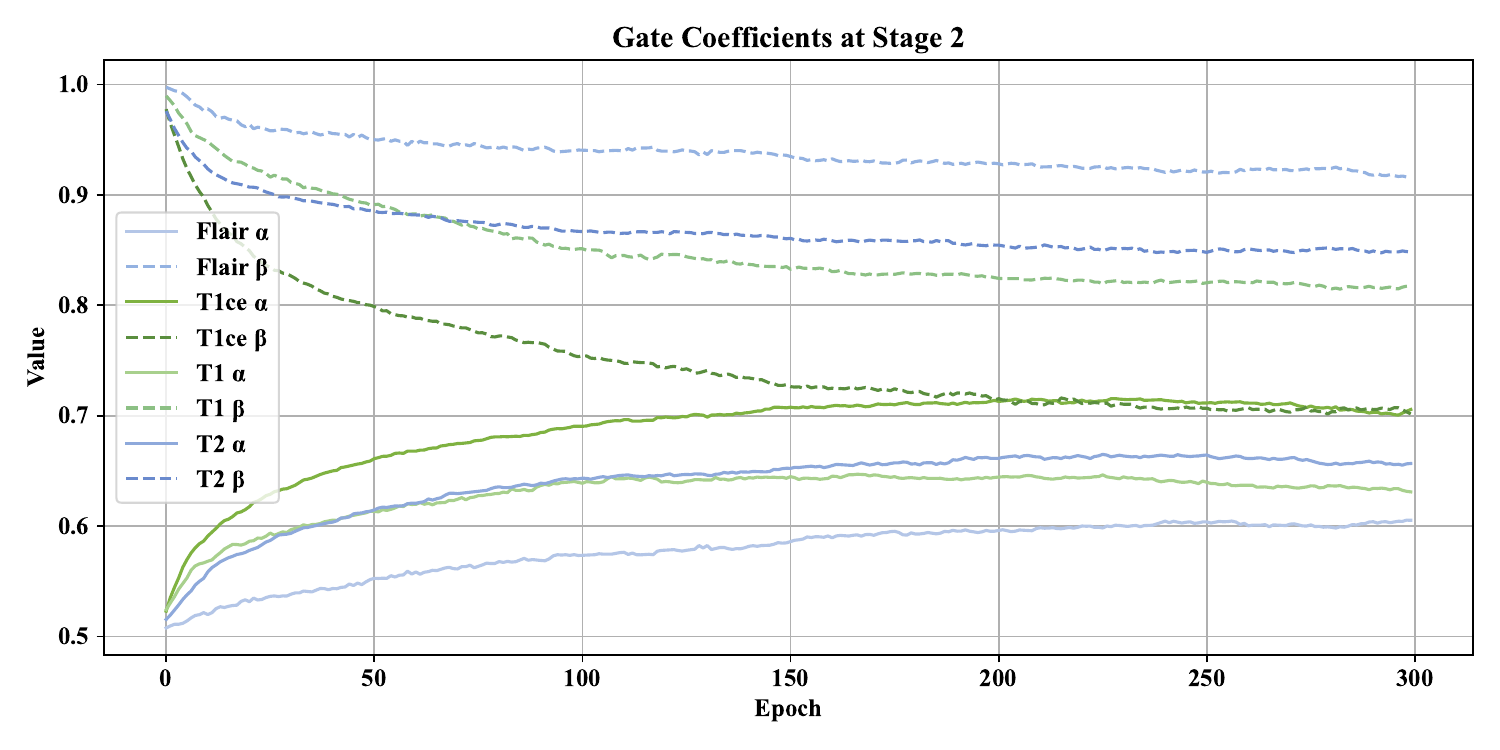} \\
      \includegraphics[width=0.45\textwidth]{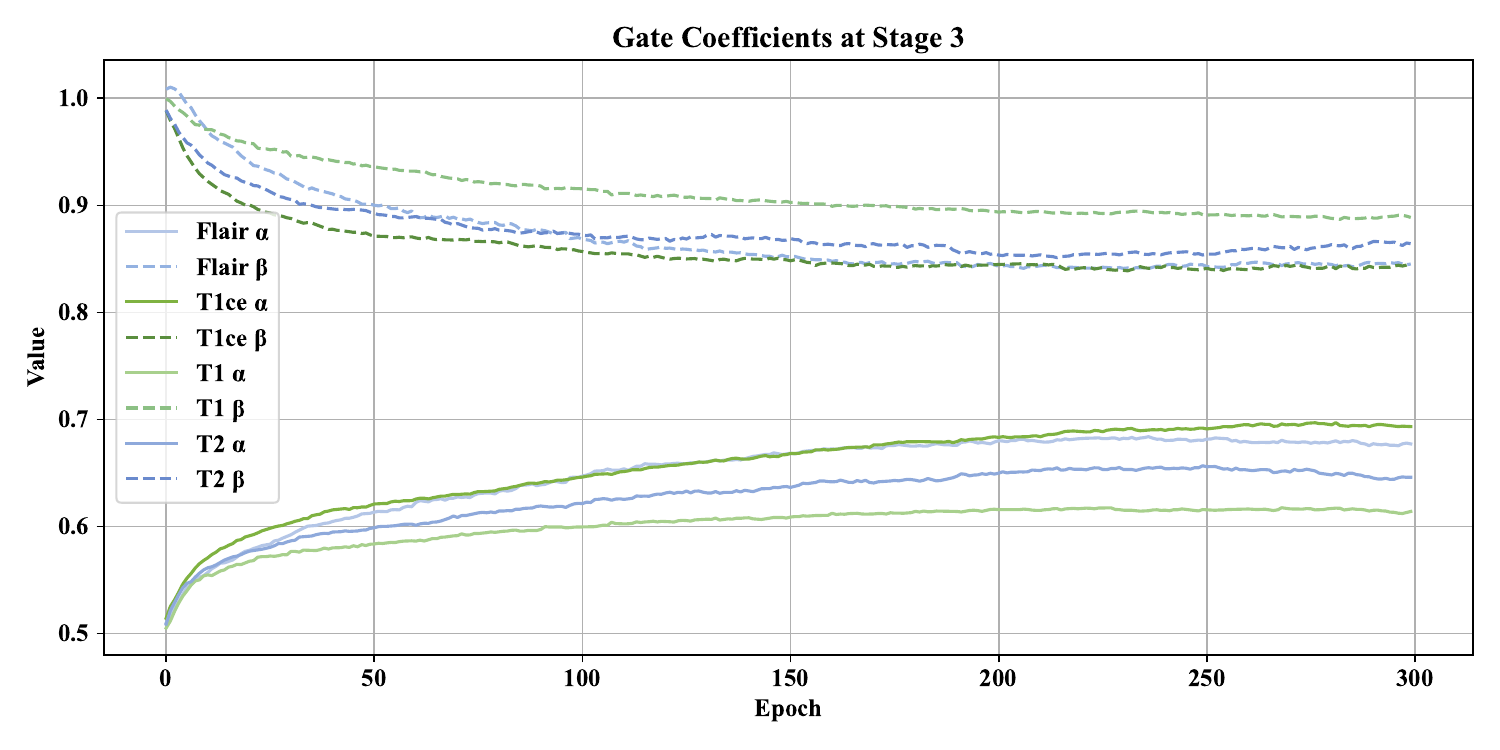} &
      \includegraphics[width=0.45\textwidth]{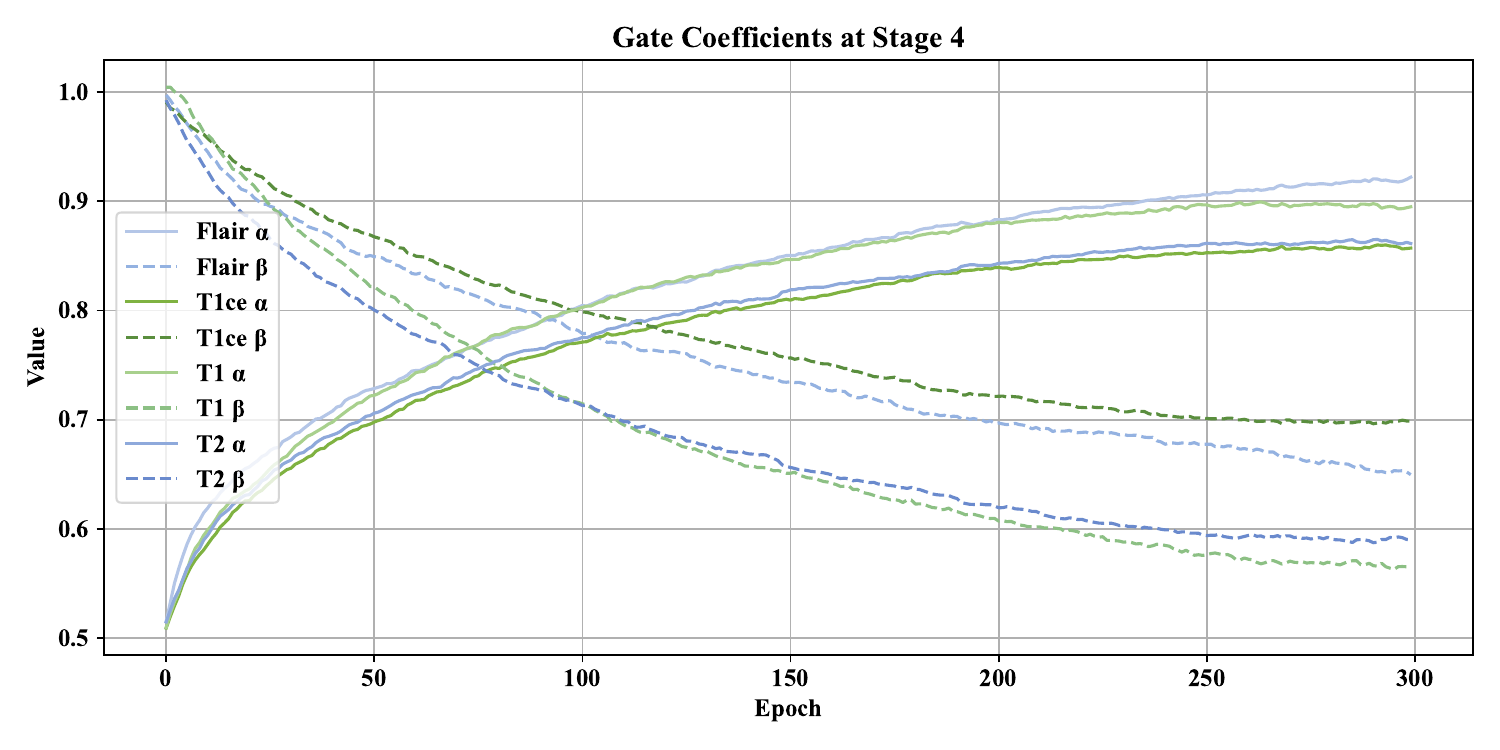}
    \end{tabular}
  }
  \vspace{0.2cm} 
  
  \subfloat[Coefficient Values Across Training Epochs\label{fig:gate_boxplots}]{%
    \includegraphics[width=0.46\textwidth,trim=0 0 0 48pt,clip]{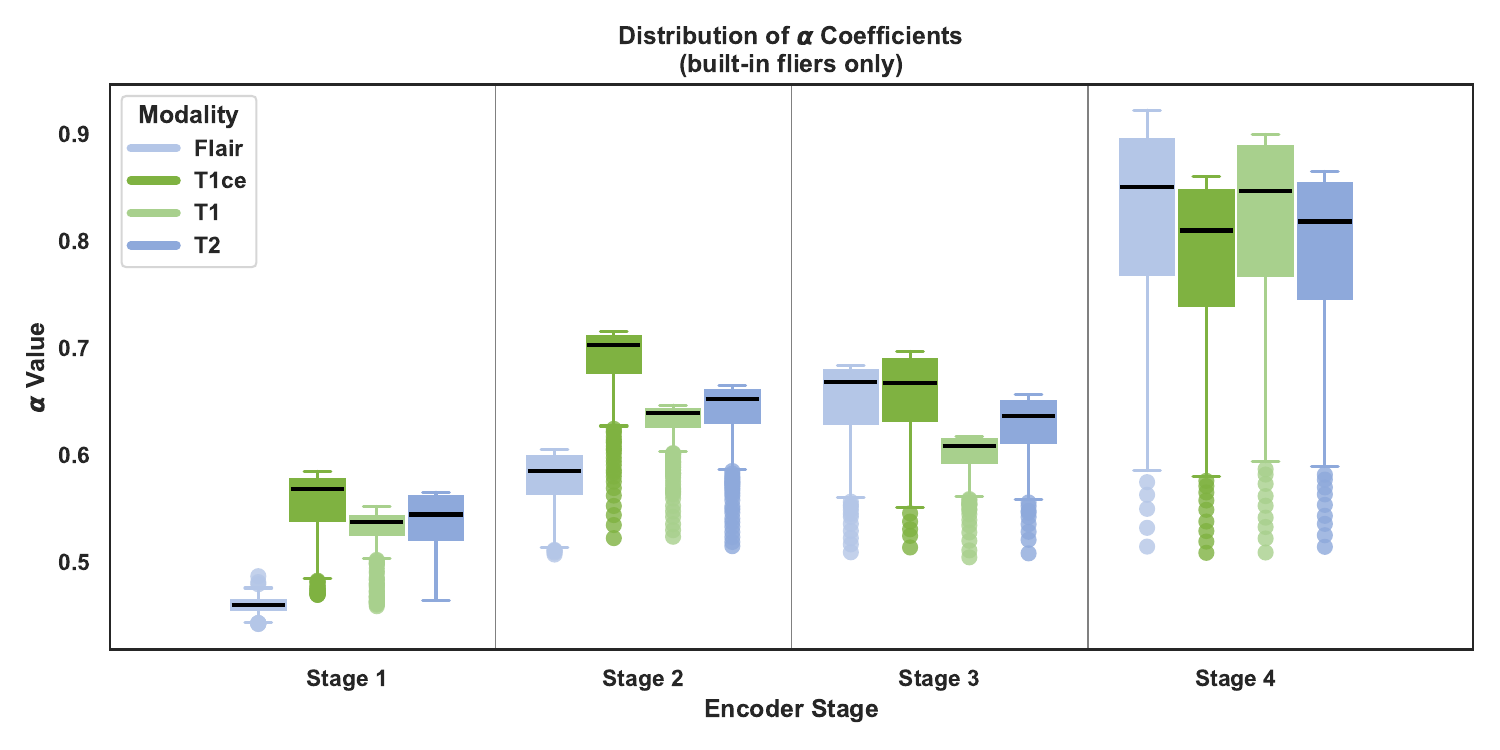}%
    \includegraphics[width=0.46\textwidth,trim=0 0 0 48pt,clip]{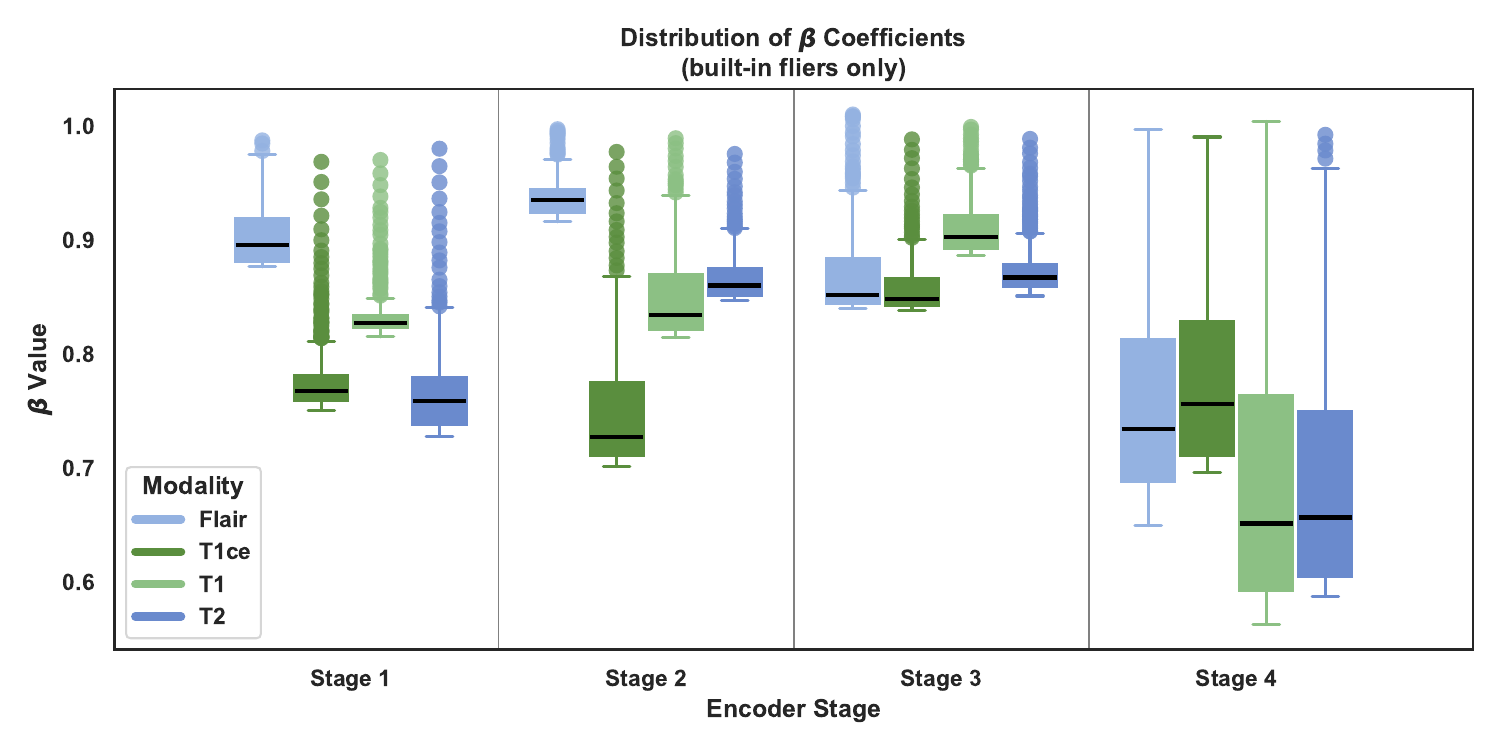}%
  }
  
  \subfloat[Final Learned Gate Heatmaps\label{fig:gate_heatmaps}]{%
    \includegraphics[width=0.46\textwidth]{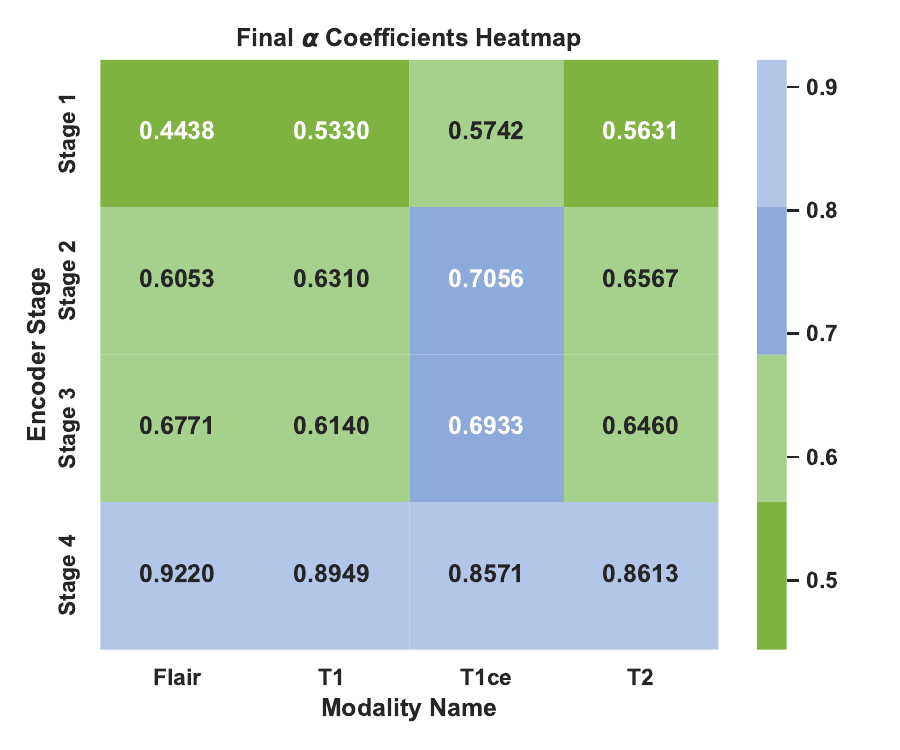}%
    \includegraphics[width=0.46\textwidth]{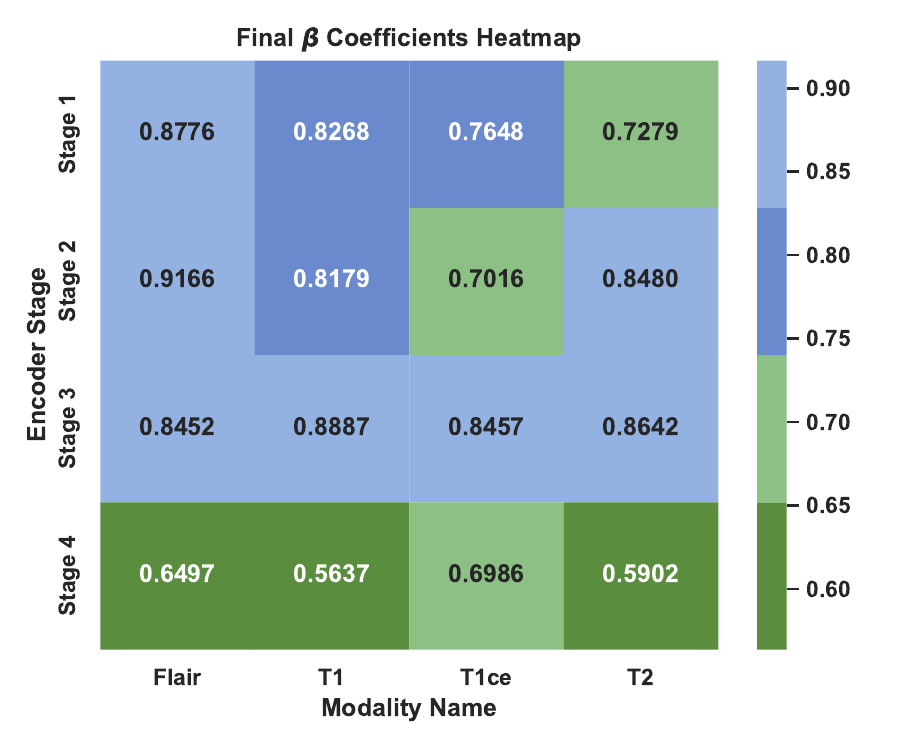}%
  }
  \caption{Learned gate coefficients for RFNet+TG-BWSA over 300 training epochs.
  \textbf{(a)} Coefficient trajectories for each modality and encoder stage.
  \textbf{(b)} Distributions of values recorded across training epochs.
  \textbf{(c)} Final coefficients after epoch 300. $\alpha$ weights the adapted stream and $\beta$ weights the original stream.}
  \label{fig:gate_analysis}
\end{figure*}

We use RFNet on the BraTS 2023 local validation set for the component and missing-modality analyses. We also examine gate initialization, regularization, training loss, and the learned gate coefficients (Figs.~\ref{fig:training_loss} and~\ref{fig:gate_analysis}).

\subsubsection{Component-wise evaluation of BWSA and its gated extension}
\label{subsec:component_ablation}
Table~\ref{tab:ablation_adapter_full} compares fixed and learned weights with changes to the main adapter components over three runs. The no-residual variant removes the direct original-feature path. In the RFNet ablation, BWSA scores 0.8955 mean Dice with fixed weights and 0.8982 with learned gates, a difference of 0.0027. Removing the original-feature path lowers the mean to 0.8944. PMA uses four independent adapters with the same internal operations and a similar total parameter count, but does not combine modality features within an adapter. BWSA with learned gates scores 0.8982, compared with 0.8947 for PMA under this RFNet protocol.

\begin{table}[!t]
\centering
\caption{Component-wise evaluation of adapter variants on BraTS 2023 using RFNet (mean$\pm$standard deviation over three runs). Best results are in \textbf{bold}, and second-best results are \underline{underlined}.}
\label{tab:ablation_adapter_full}
\scriptsize
\setlength{\tabcolsep}{2pt}
\resizebox{\columnwidth}{!}{%
\begin{tabular}{lcccc}
\toprule
\textbf{Configuration} & \textbf{WT} & \textbf{TC} & \textbf{ET} & \textbf{Mean} \\
\midrule
\textbf{TG-BWSA (learned gates)}
 & \textbf{0.9293}{\scriptsize$\pm$0.0009}
 & \textbf{0.9007}{\scriptsize$\pm$0.0011}
 & \textbf{0.8647}{\scriptsize$\pm$0.0008}
 & \textbf{0.8982} \\
\textbf{BWSA (fixed weights)}
 & 0.9261{\scriptsize$\pm$0.0015}
 & 0.8990{\scriptsize$\pm$0.0002}
 & \underline{0.8613}{\scriptsize$\pm$0.0015}
 & \underline{0.8955} \\
\textbf{Per-Modality Adapters (PMA)}
 & 0.9250{\scriptsize$\pm$0.0018}
 & 0.8985{\scriptsize$\pm$0.0009}
 & 0.8605{\scriptsize$\pm$0.0012}
 & 0.8947 \\
\textbf{No Residual Path}
 & 0.9230{\scriptsize$\pm$0.0025}
 & \underline{0.9001}{\scriptsize$\pm$0.0018}
 & 0.8601{\scriptsize$\pm$0.0020}
 & 0.8944 \\
\textbf{Baseline (RFNet)} 
 & 0.9244{\scriptsize$\pm$0.0022}
 & 0.8976{\scriptsize$\pm$0.0023}
 & 0.8603{\scriptsize$\pm$0.0019}
 & 0.8941 \\
\midrule
w/o CMCR (no channel recalibration) 
 & \underline{0.9265}{\scriptsize$\pm$0.0024}
 & 0.8964{\scriptsize$\pm$0.0020}
 & 0.8562{\scriptsize$\pm$0.0022}
 & 0.8930 \\
\makecell[l]{w/o Depthwise Separable\\Convolution}
 & 0.9251{\scriptsize$\pm$0.0027}
 & 0.8955{\scriptsize$\pm$0.0022}
 & 0.8547{\scriptsize$\pm$0.0023}
 & 0.8918 \\
Basic Adapter (1$\times$1 conv only) 
 & 0.9175{\scriptsize$\pm$0.0031}
 & 0.8870{\scriptsize$\pm$0.0026}
 & 0.8394{\scriptsize$\pm$0.0028}
 & 0.8813 \\
\bottomrule
\end{tabular}%
}
\end{table}

\subsubsection{Gate initialization and regularization}
We evaluated five combinations of $\alpha_0$, $\beta_0$, and $\lambda$ (Table~\ref{tab:sensitivity}). Across these settings, WT Dice ranges from 0.9263 to 0.9293, TC from 0.8964 to 0.9007, and ET from 0.8598 to 0.8647. The narrow ranges show limited sensitivity to the five tested settings.

\begin{table}[h!]
\centering
\caption{Effect of gate initialization and regularization on BraTS 2023 with RFNet.}
\label{tab:sensitivity}
\begin{tabular}{cccccc}
\toprule
$\alpha_0$ & $\beta_0$ & $\lambda$ & \textbf{WT} & \textbf{TC} & \textbf{ET} \\
\midrule
\textbf{0.5} & \textbf{1.0} & \textbf{1e-4} & \textbf{0.9293} & \textbf{0.9007} & \textbf{0.8647} \\
0.1 & 1.0 & 1e-4 & 0.9272 & 0.8991 & 0.8614 \\
0.5 & 0.5 & 1e-4 & 0.9269 & 0.8970 & 0.8598 \\
0.5 & 1.0 & 1e-5 & 0.9280 & 0.9001 & 0.8631 \\
0.5 & 1.0 & 1e-3 & 0.9263 & 0.8964 & 0.8602 \\
\bottomrule
\end{tabular}
\end{table}

\subsubsection{Performance with Missing Modalities}
\label{subsec:missing_modality}

\noindent\textbf{Ablation Protocol:}
At test time, a missing modality is simulated by zeroing its input volume while keeping standard preprocessing for the others. Inference is run in evaluation mode to prevent normalization updates.

\noindent\textbf{Results:}
We evaluate RFNet with and without BWSA, using either fixed weights or learned gates, over the 14 incomplete subsets of \{T1ce, T1, FLAIR, T2\}. The input containing all four modalities is reported separately with the complete input results. Table~\ref{tab:missing_summary} averages only the incomplete input configurations. Both forms of BWSA improve mean WT, TC, and ET Dice over RFNet. Fixed weights give higher TC Dice (0.796 vs.\ 0.779), while learned weights give slightly higher WT (0.834 vs.\ 0.825) and ET (0.629 vs.\ 0.627) Dice.

\begin{table}[!tbh]
\centering
\caption{Mean Dice across the 14 incomplete-input modality configurations on BraTS 2023. The complete-input case is excluded and reported with the full-modality results. Differences in the text are calculated from unrounded means.}
\label{tab:missing_summary}
\begin{tabular*}{\linewidth}{@{\extracolsep\fill}lccc}
\toprule
\textbf{Region} & \textbf{RFNet} & \textbf{+BWSA} & \textbf{+TG-BWSA} \\
\midrule
WT & 0.754 & 0.825 & \textbf{0.834} \\
TC & 0.531 & \textbf{0.796} & 0.779 \\
ET & 0.430 & 0.627 & \textbf{0.629} \\
\bottomrule
\end{tabular*}
\end{table}

\subsubsection{Training Loss and Gate Analysis}
\label{subsec:gate_analysis}

Figure~\ref{fig:training_loss} compares the recorded RFNet and RFNet+TG-BWSA training losses. The TG-BWSA curve is lower over most of the plotted epochs, but training loss alone does not establish convergence rate or validation performance.

Figure~\ref{fig:gate_analysis} shows the trajectories of $\alpha$ and $\beta$ over 300 training epochs, distributions of their recorded values across those epochs, and their final values for each modality and stage. The boxplot medians summarize the training trajectories and need not equal the final coefficients. After training, the coefficients are fixed model parameters and do not adapt to individual test inputs.

For FLAIR at Stage~4, $\alpha$ increases from its initialization near 0.5 to 0.9220. At Stage~2, the final $\alpha$ and $\beta$ values for T1ce are close (0.7056 and 0.7016). The final coefficients differ across modalities and stages. They weight the adapted and original streams, but do not establish a link between an individual gate and a particular tumor region.

\section{Discussion}
\label{sec:discussion}

BWSA increases mean Dice over the unmodified model across the four evaluated backbones in local and online validation. Learning the weights does not improve every metric. On BraTS 2023 online validation, the gated form has higher average Dice on all four backbones, but fixed weights give RFNet a lower average HD95 (14.37 vs.\ 16.34\,mm; Table~\ref{tab:table5}). Under incomplete RFNet inputs, fixed weights give higher mean TC Dice, whereas learned weights give higher WT and ET Dice (Table~\ref{tab:missing_summary}). For MEUNet, learning the weights raises average online Dice from 0.8303 to 0.8460. Both forms of BWSA improved average Dice and HD95 over the unmodified backbones on BraTS 2023 online validation. Their parameter counts are nearly identical. For a given backbone, the choice between fixed blending and learned weights should be guided by validation performance on the metric of interest.

The gate coefficients are fixed after training and cannot adjust to the reliability of an individual test image. Their trajectories do not explain the observed differences between tumor regions. In the RFNet ablation, TG-BWSA has higher mean Dice than PMA with a similar parameter count (0.8982 vs.\ 0.8947; Table~\ref{tab:ablation_adapter_full}). The measured latency, memory, and GFLOPs apply to RFNet on an NVIDIA A6000 (Table~\ref{tab:inference_efficiency}); the cost for the other backbones was not measured.

For the lesion-wise online metrics, RFNet+TG-BWSA has higher average Dice (0.852 vs.\ 0.848) and lower average HD95 (20.61 vs.\ 23.40\,mm) than RFNet+BWSA (Table~\ref{tab:table6}). Its average lesion-wise Dice remains below the three highest-scoring teams listed in that table (0.876--0.888). The compared submissions use different training and inference pipelines, so these scores provide benchmark context rather than an isolated comparison of architectures.

\subsection*{Limitations and Future Work}
\label{subsec:limitations}

The evaluation uses BraTS 2020 and BraTS 2023 without an external cohort. Each online score comes from one of three runs selected by local validation Dice, so variation in online scores across seeds was not measured. Missing modalities are simulated by zeroing complete input volumes, and modality dropout is not used during training. The LoRA comparison differs in parameter count and module placement, so it does not measure efficiency per parameter. Future studies can test scanner and protocol shifts, partial image corruption, and modality dropout during training.

\section{Conclusion}

BWSA lets separate modality encoders combine features through one shared adapter between stages. With fixed weights, it adds the adapted features to the original features. TG-BWSA extends the same adapter by learning those weights for each modality and stage. We evaluated both forms with four backbones on BraTS 2020 and BraTS 2023. On the official BraTS 2023 online validation set, BWSA increased average region-wise Dice by 1.3--7.6 percentage points relative to the unmodified backbones. The gated extension increased it by 2.8--7.7 points. Both forms increased parameter counts by approximately 0.85--3.73\% relative to the unmodified backbones. Both forms reduced average HD95 relative to the corresponding unmodified backbones. The choice between fixed blending and learned weights should be guided by validation results for the chosen backbone and evaluation metric.

\section*{Data availability}
The BraTS 2020 and BraTS 2023 datasets analyzed in this study are available from the BraTS challenge repositories, subject to the respective access and data-use conditions. BraTS~2023 is distributed via Synapse.

\section*{Ethics statement}
This secondary analysis used publicly available and deidentified BraTS 2020 and BraTS 2023 data. The authors recruited no participants, collected no new human subject data, and accessed no identifiable information. Therefore, no additional institutional review board approval or informed consent was required. Original data collection and release followed the ethics and data use procedures of the contributing institutions and the BraTS organizers.

\section*{Code availability}
Code and experiment scripts are available at \url{https://github.com/milad67/TG-BWSA}.

\section*{Author contributions}
M.R. contributed to conceptualization, methods, software, analysis, and writing. M.M. contributed to methods and design. E.K. contributed to interpretation and revision. M.S. supervised the study. All authors approved the manuscript.

\EOD

\end{document}